\documentclass{article} 
\usepackage{template}

\usepackage{microtype}
\usepackage[colorlinks=true]{hyperref}
\usepackage{url}
\usepackage{booktabs}
\usepackage{algorithm}
\usepackage[noend]{algpseudocode}
\usepackage{float}
\usepackage{amssymb}
\usepackage[pdftex]{graphicx} 
\usepackage{subcaption}
\usepackage{amsmath}
\usepackage{multirow}

\title{Modeling Real-Time Interactive Conversations as Timed \\ Diarized Transcripts}

\author{Garrett Tanzer\thanks{~denotes equal contribution; order chosen by a simulated conversation between the first authors. Correspondence to \texttt{gahdritz@g.harvard.edu}. See our code at \repolink.} \\
Google
\And
Gustaf Ahdritz$^*$ \\
Harvard University
\And
Luke Melas-Kyriazi \\
Oxford University
}

\algblockdefx{Try}{EndTry}[1]{\textbf{try} #1}{\vspace{-\baselineskip}\vspace{0.15em}}
\algblockdefx{Catch}{EndCatch}[1]{\textbf{catch} #1}{\vspace{-\baselineskip}\vspace{0.15em}}

\newcommand{\repolink}{\href{https://github.com/gahdritz/rtic}{this link}}
\newcommand{\messengercodelink}{\href{https://github.com/gahdritz/rtic}{this link}}
\newcommand{\speechcodelink}{\href{https://github.com/gahdritz/rtic}{this link}}
\newcommand{\ninetynine}{28}
\newcommand{\ninetyninepointnine}{75}
\newcommand{\optimalninetynine}{28}
\newcommand{\chattokens}{20.2M}
\newcommand{\scotustokens}{40.3M}
\newcommand{\speechninetynine}{36}
\newcommand{\speechninetyninepointnine}{45}
\newcommand{\optimalspeechninetynine}{22}
\newcommand{\optimalspeechninetyninepointnine}{30}

\colmpreprint
\begin{document}

\maketitle

\vspace{-2mm}
\begin{abstract}
Chatbots built upon language models have exploded in popularity, but they have largely been limited to synchronous, turn-by-turn dialogues. In this paper we present a simple yet general method to simulate real-time interactive conversations using pretrained text-only language models, by modeling \textit{timed diarized transcripts} and decoding them with \textit{causal rejection sampling}. We demonstrate the promise of this method with two case studies: instant messenger dialogues and spoken conversations, which require generation at about 30 tok/s and 20 tok/s respectively to maintain real-time interactivity. These capabilities can be added into language models using relatively little data and run on commodity hardware.
\end{abstract}

\section{Introduction}
\label{sec:introduction}

Chatbots built upon language models have exploded in popularity, but their interaction model is extremely limited: the user and the system take turns writing messages, where the system waits until the user finishes their message to respond then responds instantly and uninterruptibly. Extensions to support audio have used speech to text and text to speech to eliminate the need for typing and reading the screen~\citep{chatgptaudio}, but the constraints of the interaction model have remained the same.

In this paper we present a simple yet general method to simulate real-time interactive conversations using pretrained text-only language models, namely: model \textit{timed diarized transcripts}---i.e., sequences of [timestamp, speaker id, message]---at the desired granularity, and then decode these transcripts with \textit{causal rejection sampling}---i.e., sample a continuation that will be finalized at the predicted timestamp, and if there is intervening user input before the timestamp, reject the planned continuation (to the extent that its probability under the model has changed) and resample a new one. This method is naturally sparse over time and number of speakers, scaling computation with the amount of content being actively produced at each moment in time.

We demonstrate the promise of this method with case studies in two domains. First, we use the instant messenger chat history between the first authors to train a real-time interactive asynchronous text dialogue model. Second, we use public speech datasets with diarized transcripts to train a real-time spoken conversation model, cascaded through word-level speech to text and text to speech models. Here there is an additional complication in that real-time streaming speech to text systems are unstable, i.e., predictions may change in light of future context. We address this with \textit{retconning}, i.e., revising the user's input history but keeping any already finalized system outputs.

We evaluate these embodiments of our method with respect to performance (properties of the control token format and of our proof-of-concept implementation) and quality (test perplexity, offline human ratings, and online human ratings)---across finetuned models from 160M to 12B parameters. For the offline human rating setting only, we also use long in-context learning to test larger pretrained models available by API. In order to maintain real-time interactivity, generation needs to be about {\optimalninetynine} tokens per second for the instant messenger use case and {\optimalspeechninetynine} tok/s for spoken conversations, which are easy to achieve on a single A100 at our model scales. We find that, predictably, better pretrained models lead to better results, though there is still obvious room for improvement with dataset/model scale.

We publicly release our code (and some demo videos) at \repolink. We hope that these proofs of concept spark the imagination and show that language models can easily be adapted to new real-time interaction modes.

\section{Method}
\label{sec:method}

We model \textit{timed diarized transcripts} using causally masked (decoder-only) language models. Given a sequence of events $e_i$, where each event $e_i$ consists of a timestamp $t_i$ (\textit{timed}), a speaker id $s_i$ (\textit{diarized}), and a message $m_i$ (\textit{transcript}), we model $p(e_i | e_1, ..., e_{i-1})$. In practice, this function decomposes into $p(t_i | e_1, ..., e_{i-1})$, $p(s_i | e_1, ..., e_{i-1}, t_i)$, and $p(m_i | e_1, ..., e_{i-1}, t_i, s_i)$, or even more granular distributions if these components are represented as multiple tokens. By modeling events sparsely over time, we are able to sample transcripts with computation proportional to the number/complexity of the events, rather than the time duration.

In order to make this model interactive, we use \textit{causal rejection sampling}. We pick a particular speaker id $S$ to represent the user and sample candidates $\hat{e}_i \sim p(e_i | e_1, ..., e_{i-1})$, where we interpret the timestamps $t$ within these events with respect to the current real time. If an input from the user $(S, T, M)$ interrupts before the timestamp $\hat{t}_i$ is reached, we reject the candidate $\hat{e}_i$ and sample a new candidate $\hat{e}_{i+1} \sim p(e_{i+1} | e_1, ..., e_{i-1}, e_i=(S, T, M))$. If no such interruption occurs before $\hat{t}_i$, there are two possibilities: If the speaker id $\hat{s}_i$ within $\hat{e}_i$ is not $S$, we accept the message candidate $\hat{m_i}$, emit it to the user, then sample $\hat{e}_{i+1}$, etc. If $\hat{s}_i$ is $S$, then we resample $\hat{e}'_i \sim p(e_i | e_1, ..., e_{i-1}, t_i \geq \hat{t}_i)$.

Because it takes some amount of time $t_{latency}$ (varying with message length) to execute the model and sample from $p(e_i | ...)$, if the user repeatedly provides input less than $t_{latency}$ before the predicted timestamps $\hat{t}_i$, the model will be starved and unable to generate any acceptable events. We provide two modifications to mitigate recomputation from user interruption:

First, we enforce a hard lower bound on the model's generation bandwidth by stipulating that if the user input comes within $t_{react}$ of $\hat{t}_i$, we accept $\hat{e}_i$ as a candidate for $\hat{e}_{i+1}$. The relationship between $t_{latency}$ and $t_{react}$ determines whether the model can maintain real-time interactivity in the worst case. We do not expect moderate $t_{react}$ to harm generation quality too much because a human reaction time of approximately 150-200 ms~\citep{humanVisualReactTime, humanAudioReactTime} should be reflected in the underlying causal structure of human training data.

Second, we reduce the average amount of recomputation by integrating speculative decoding~\citep{leviathan2023fast, chen2023accelerating}. Rather than discard the candidate $\hat{e}_i$ unconditionally upon user interruption, we treat it as a draft for the new generation, rejecting and resampling based on the closeness of $p(e_i = \hat{e}_i | e_1, ..., e_{i-1}, t_i \geq T)$ and $p(e_{i+1} = \hat{e}_i | e_1, ..., e_{i-1}, e_i=(T, S, M))$. Note that this is different from traditional speculative decoding, where a smaller model \textit{for the same distribution} drafts a candidate;\footnote{The traditional kind of speculative decoding could also be used to speed up the initial autoregressive candidate generation; we omit this for simplicity.} the use of different prompts under the same model resembles classifier-free guidance~\citep{ho2022classifierfree, sanchez2023stay}. Like with $t_{react}$, we expect this to work to the extent that there is a looseness in the causal dependencies of nearby messages from different parties.\footnote{You can also trade off between potentially wasted computation and interactivity by sampling the timestamp first and waiting until it approaches to generate the rest of the message, vs. sampling multiple sequential event candidates ahead of time.}

See Algorithm~\ref{alg:basic-alg} for a formal description of causal rejection sampling (speculative decoding omitted for clarity; see Appendix~\ref{app:speculative-algorithm} for the full version), or see our code at \messengercodelink.

\begin{figure}[t]
\centering
\begin{minipage}{0.8\linewidth}
\begin{algorithm}[H]
\caption{Causal rejection sampling (without speculative decoding)}\label{alg:basic-alg}
\begin{algorithmic}
\State $i \gets 0$ \Comment{current event index}
\State $e \gets []$ \Comment{event history}
\State $c \gets (\varnothing, \varnothing, \varnothing)$ \Comment{candidate for the next message}
\While {true}
\State $i \gets i + 1$
\Try
\State $( \hat{t}, \hat{s}, \hat{m}) \gets c$
\If{$\hat{t}$ is $\varnothing$}
\State $c \gets (\hat{t}, \hat{s}, \hat{m}) \sim p(e_i | e_1, ..., e_{i-1}, t_i \geq t_{cur})$
\EndIf
\State \textbf{wait} until $\hat{t}$
\State $t_{cur} \gets \hat{t}$
\If{$\hat{s}$ is $S$}
\State $c \gets (\varnothing, \varnothing, \varnothing)$
\State $i \gets i-1$
\State \textbf{continue}
\EndIf
\EndTry
\Catch{user input $(T, S, M)$}
\State $e_i \gets (T, S, M)$
\State $t_{cur} \gets T$
\State $(\hat{t}, \hat{s}, \hat{m}) \gets c$
\If {$\hat{s} = S$ or $\hat{t} + t_{react} < T$}
\State $c \gets (\varnothing, \varnothing, \varnothing)$
\EndIf
\State \textbf{continue}
\EndCatch
\State $e_i \gets c$
\State \textbf{emit} $c$
\State $c \gets (\varnothing, \varnothing, \varnothing)$
\EndWhile
\end{algorithmic}
\end{algorithm}
\end{minipage}
\vspace{-2mm}
\end{figure}

We now present two case studies demonstrating how this method can be applied to different domains: instant messenger dialogues and spoken conversations.

\subsection{Instant Messenger Dialogues}
\label{subsec:messenger-method}

The method as described above can be applied to instant messenger dialogues with minimal modifications. We use as our domain 9 years of instant messenger history between the first authors. This means we are not just modeling the evolution of synchronous conversations where both participants are actively engaged, but asynchronous conversations where participants may be offline and where the date/time may influence the content of the conversation. Instant messenger conversations can be highly multimodal, in particular with audio, images, and hyperlinks; we consider only text and leave multimodality to future work.

\begin{figure}
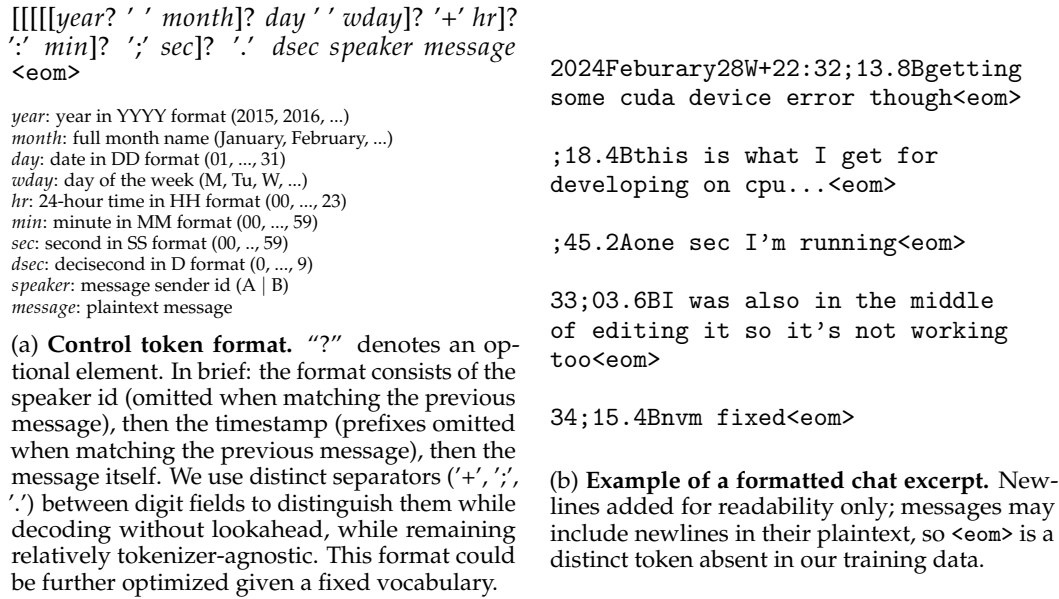

\centering
    \begin{subfigure}[b]{.48\textwidth}
        [[[[[$year$? ' ' $month$]? $day$ ' ' $wday$]? '+' $hr$]? ':' $min$]? ';' $sec$]? '.' $dsec$ $speaker$ $message$ \texttt{<eom>} \\
\hspace{0mm} \\
{
\scriptsize
$year$: year in YYYY format (2015, 2016, ...) \\
$month$: full month name (January, February, ...) \\
$day$: date in DD format (01, ..., 31) \\
$wday$: day of the week (M, Tu, W, ...) \\
$hr$: 24-hour time in HH format (00, ..., 23) \\
$min$: minute in MM format (00, ..., 59) \\
$sec$: second in SS format (00, .., 59) \\
$dsec$: decisecond in D format (0, ..., 9) \\
$speaker$: message sender id (A $|$ B) \\
$message$: plaintext message
\par
}
\caption{\textbf{Control token format.} ``?'' denotes an optional element. In brief: the format consists of the speaker id (omitted when matching the previous message), then the timestamp (prefixes omitted when matching the previous message), then the message itself. We use distinct separators ('+', ';', '.') between digit fields to distinguish them while decoding without lookahead, while remaining relatively tokenizer-agnostic. This format could be further optimized given a fixed vocabulary.}
\label{fig:messenger-format}
    \end{subfigure}
    \hspace{0.02\textwidth}
    \begin{subfigure}[b]{.48\textwidth}
        \texttt{2024Feburary28W+22:32;13.8Bgetting some cuda device error though<eom>} \\ \\
\texttt{;18.4Bthis is what I get for developing on cpu...<eom>} \\ \\
\texttt{;45.2Aone sec I'm running<eom>} \\ \\
\texttt{33;03.6BI was also in the middle of editing it so it's not working too<eom>} \\ \\
\texttt{34;15.4Bnvm fixed<eom>} \\
\caption{\textbf{Example of a formatted chat excerpt.} Newlines added for readability only; messages may include newlines in their plaintext, so \texttt{<eom>} is a distinct token absent in our training data.\vspace{3mm}}
\label{fig:messenger-example}
    \end{subfigure}
    \caption{\textbf{Formatting for the instant messenger case study.}}
\end{figure}

In the notation from above, we instantiate $t$ with the message's calendar date/time (down to decisecond granularity), $s$ with an id representing the message sender (one of the two authors), and $m$ with the message plaintext (terminated by an ``end of message'' token). As a sequence length optimization, when prefixes of the timestamp are repeated in consecutive messages, we omit them. We design the control format to be prefix-free so that it can be interpreted without lookahead while decoding; this means that control tokens can decoded in a structured way (including that time only flows forward) by appropriately filtering and renormalizing the next token vocabulary. See Figure~\ref{fig:messenger-format} for a specification of the format and Figure~\ref{fig:messenger-example} for an example of what preprocessed data looks like.

\subsection{Spoken Conversations}
\label{subsec:spoken-method}

We also apply our general method to timed diarized word-level automatic speech recognition (ASR) transcripts. By cascading input through speech-to-text and output through text-to-speech, we can simulate spoken conversations. Note that---like cascaded approaches in general---this has the obvious limitation that it bottlenecks the input and output through text, stripping away aspects of speech like tone and introducing errors from intermediate models. While there exist off-the-shelf streaming speech-to-text models that output word-level timestamps, we are not aware of any text-to-speech models (streaming or otherwise) that accept them as input: the closest is incremental text-to-speech~\citep{ma2020incremental}. This limits our ability to generate natural-sounding speech; we use word-level text to speech invoked at the specified timestamps and consider this out of scope.

There is an additional complication due to the use of streaming speech-to-text models: these models are able to achieve low latency because they output preliminary transcriptions that may change in light of future input and are only finalized some time later. This means that not only can the user's input interrupt the model's candidate generation, but the input can retroactively change after a candidate has been generated, accepted, and spoken out.

We address this with \textit{retconning}, i.e., when the speech-to-text model's prediction for the input changes, we replace the old prediction with the new one in the transcript prefix, without changing any model generations that were accepted after that point. More formally, if we have sampled $\hat{e}_{j} \sim p(e_{j} | e_1, ..., e_{i}, ..., e_{j-1})$ and the user interrupts with a revision $e'_{i}$, we reject $\hat{e}_{j}$ (subject to the $t_{react}$ window and speculation described above) and resample $\hat{e}'_{j} \sim p(e_{j} | e_1, ..., e'_{i}, ..., e_{j-1})$. This should not have a significant impact on either performance or quality, since processing $n$ tokens in parallel is much faster than $n$ tokens sequentially, and because humans also reinterpret what they've already heard in light of new speech (which should be reflected in ground truth causal structure). See Appendix~\ref{app:spoken-algorithm} for a more formal description of causal rejection sampling with retconning, or see our code at \speechcodelink.


We use as our dataset 1000 hours of oral arguments before the U.S. Supreme Court~\citep{oyez,scotusrepo}. Court oral arguments are an interesting domain because they have many participants ($\sim$10 per transcript) and are information dense, though they have longer conversation turns and fewer interruptions than typical conversations.

In the formal language from Section~\ref{sec:method}, we instantiate $t$ with the word's start timestamp modulo 10 seconds\footnote{This compromise reduces the number of tokens, at the expense of being able to model more than 10 seconds of silence.} (down to centisecond granularity), $s$ with an opaque identifier representing the speaker, and $m$ with the word plaintext (terminated by an ``end of message'' token). 
We omit the speaker id in repeated spans. See Figure~\ref{fig:speech-format} for a more complete description of the format and Figure~\ref{fig:speech-example} for an example of what preprocessed data looks like.\footnote{Note that for generality, the duration of each word should probably also be modelled. We omit it here because it would be discarded in our word-level text to speech step anyway.}

\begin{figure}
\centering
    \begin{subfigure}[b]{.48\textwidth}
        $sec$ $dsec$ $csec$ $speaker$ $word$ \texttt{<eom>} \\
\hspace{0mm} \\
{
\scriptsize
$sec$: ones place of the timestamp in seconds (0, .., 9) \\
$dsec$: tenths place of the timestamp (0, ..., 9) \\
$csec$: hundredths place of the timestamp (0, ..., 9) \\
$speaker$: speaker id (A, B, ...) \\
$word$: plaintext word
\par
}
\caption{\textbf{Control token format.} ``?'' denotes an optional element. In brief: the format consists of the speaker id (omitted when matching the previous message), then the timestamp (prefixes omitted when matching the previous message), then the message itself.}
\label{fig:speech-format}
    \end{subfigure}
    \hspace{0.02\textwidth}
    \begin{subfigure}[b]{.48\textwidth}
\texttt{055Aknock} \\
\texttt{079Aknock} \\
\texttt{154Bwho's} \\
\texttt{186Bthere} \\
\texttt{252Ainterrupting} \\
\texttt{316Acow} \\
\texttt{377Binterrupting} \\
\texttt{443Bcow} \\
\texttt{448Amoo} \\
\texttt{473Bwho} \\
\caption{\textbf{Example of a formatted word-level transcript} (out of domain). Newline serves as \texttt{<eom>}.}
\label{fig:speech-example}
    \end{subfigure}
    \caption{\textbf{Formatting for the spoken conversation case study.}}
\end{figure}

\section{Evaluation}
\label{sec:evaluation}

For both case studies we evaluate performance and quality. We finetune the following models: Pythia 160M, 1.4B, \& 12B~\citep{biderman2023pythia}, Gemma 2B~\citep{gemma}, and Llama 2 7B~\citep{touvron2023llama}; see Appendix~\ref{app:finetuning} for details. 
Where possible, we also compare with in-context learning using state-of-the-art commercial language models: Claude 3 Sonnet~\citep{claude3} and GPT-4 Turbo~\citep{gpt4turbo}. See Appendix~\ref{app:in-context-learning} for details.

For performance, we report:
\begin{itemize}
    \setlength\itemsep{0.075em}
    \item \textit{generation bandwidth} in tokens/second required to maintain real-time interactivity, scored on historical data
    \item \textit{control token overhead ratio}, scored on historical data
    \item \textit{speculation acceptance rate} as an average number and fraction of draft tokens, scored on historical data
    \item performance properties for the proof of concept implementation
\end{itemize}

For quality, we report: 
\begin{itemize}
    \setlength\itemsep{0.075em}
    \item \textit{document-level negative log likelihood (NLL)} on the held out test set (rather than token-level perplexity, to make comparisons meaningful across tokenizers)
    \item \textit{offline human ratings}, i.e., a human ranks conversations that were generated by continuing a prefix from the test set noninteractively
    \item \textit{online human ratings}, i.e., a human interacts with each model given a conversation prefix from the test set, and then ranks them
    \item statistics about the distribution of predicted time gaps, compared to historical data
\end{itemize}

For human rating settings, we use the same prefixes of 64 messages ($\sim$1024 tokens) across all models. For the offline ratings, we also compare with the ground truth continuation. 
Note that while context lengths have recently made massive strides (128K for GPT-4 Turbo~\citep{gpt4turbo}, $>$1M for Claude 3~\cite{claude3}, and $>$10M for Gemini 1.5~\citep{gemini1p5}), they are still not long enough to fit our training sets ({\chattokens} tokens of messenger history and {\scotustokens} tokens of oral arguments) and usage is subject to rate limits. We therefore use only the most recent 16K tokens of history as context.

One of the first authors prepared the test harness; the other served as the rater. The human evaluation scores range from 0 to 6, where 0 is nonsensical and 6 is indistinguishable from real. These scores should only be used to judge relative quality and not quality in absolute.

\subsection{Instant Messenger Dialogues}
\label{subsec:messenger-eval}

As our dataset we use 9 years of instant messenger conversation history between the first authors, totaling 37,649,697 characters across 1,393,508 text-based messages (we exclude messages from other modalities). 
We use the first 95\% of the messages as the train set, the next 2.5\% as a validation set, and the last 2.5\% as a test set.

\subsubsection{Performance}
\label{subsubsec:messenger-perf}

\begin{figure}[t]
\includegraphics[width=\textwidth]{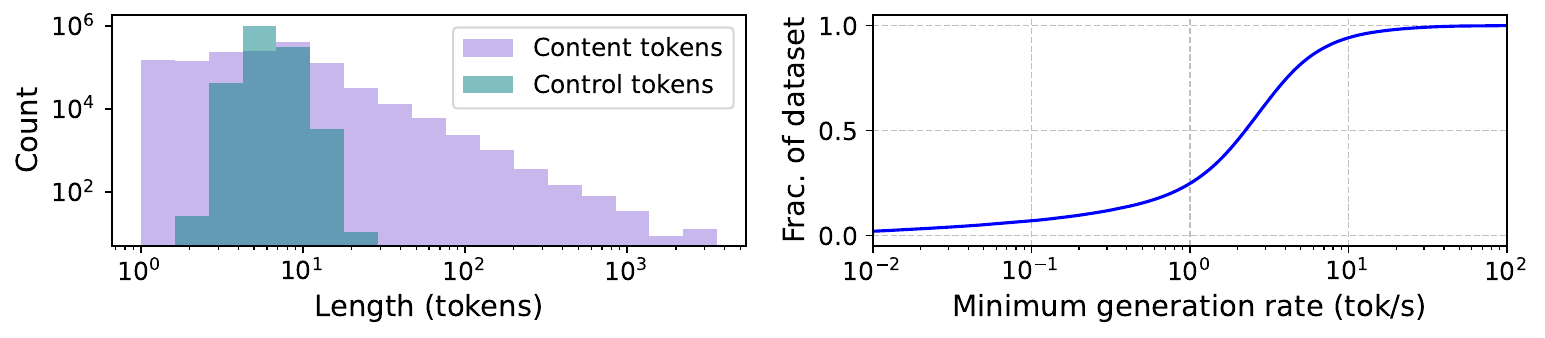}
\includegraphics[width=\textwidth]{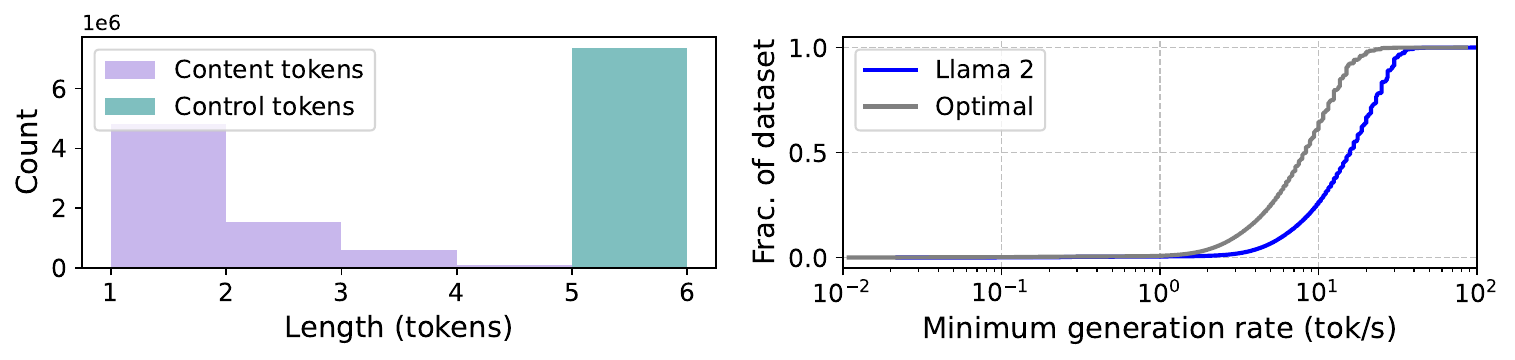}
\vspace{-7mm}
\caption{\small \textbf{Statistics about the overhead of our control formats for instant messenger dialogues (\textit{top}) and spoken conversations (\textit{bottom}), and the requirements to maintain real-time interactivity.} \textit{Left:} Lengths (in Llama 2 tokens) of plaintext messages vs. control tokens for examples in the training set. \textit{Right:} Fractions of the messages in the ground-truth dataset, including control tokens, that could be generated in real time for a given minimum generation rate, in tokens per second (again using the Llama 2 tokenizer). A message $m$ can be generated in real time if it can be generated in the time between the latest message outside of a short reaction window ($t_{react} = 200$ms) immediately before $m$, and $m$ itself. (We assume that for small $n$, the increase in cost for passing $n$ tokens through the network in parallel vs. 1 token is negligible, i.e. we are primarily modeling the cost of generating system responses, not ingesting user inputs.) For spoken conversations, we include performance figures for an optimized tokenizer which treats uses a single token for 3-digit timestamps.}
\label{fig:joint-performance-reqs}
\end{figure}

See Figure~\ref{fig:joint-performance-reqs} for details on the performance properties of our instant messenger control format. The highlights are: With $t_{react}$ = 200ms, the 99th percentile generation bandwidth required to maintain real-time interactivity is {\ninetynine} tok/s, and the 99.9th percentile is {\ninetyninepointnine} tok/s. This range is largely pathological cases like long pasted text. On average, the control-formatted token length is 3.2x the plaintext length (median 2.4x); speculative sampling saves an additional 11.02 draft tokens (69.5\% of tokens) per interruption in Llama 2.

In terms of our prototype: We interact with an A100 40GB server executing unquantized off-the-shelf model inference over \texttt{ssh}; this is more than sufficient to maintain real-time interactivity with all of our finetuned models. Communication latency is negligible, and the model checks for interruptions after generating each token (i.e., $\frac{1}{\text{\# tok/s}}$ latency).

\subsubsection{Quality}
\label{subsubsec:messenger-quality}

See Table~\ref{tab:joint-quality} for instant messenger quality results across models; see Appendix~\ref{app:qualitative-examples} for qualitative examples. The trends are unsurprising: better pretrained models achieve better perplexity and better human ratings, though still substantially worse than the ground truth. One exception is that API-based models with in-context learning mimic style worse than finetuned models, and sometimes fail completely due to refusals. We now describe some qualitative observations:

\begin{table}
\centering
\small
\begin{tabular}{l|c|cc|cc}
\toprule
& \textbf{NLL} ($\downarrow$) & \multicolumn{2}{c|}{\textbf{Offline Human Ratings} ($\uparrow$)} & \multicolumn{2}{c}{\textbf{Online Human Ratings} ($\uparrow$)} \\
\textbf{Instant messenger} & & Consistency & Fidelity & Consistency & Fidelity\\
\midrule
\texttt{Pythia 160M (ft)} & 3181 & 1.45 & 3.00 & 1.4 & 2.6 \\
\texttt{Pythia 1.4B (ft)} & 2397 & 2.55 & 3.65 & 3.4 & \textbf{4.8} \\
\texttt{Pythia 12B (ft)} & 2305 & 2.90 & 3.70 & 3.0 & 3.0 \\
\texttt{Gemma 2B (ft)} & 2376 & 2.95 & 3.65 & 2.8 & 3.2 \\
\texttt{Llama 2 7B (ft)} & \textbf{2179} & 3.90 & \textbf{4.40} & \textbf{3.8} & 4.2 \\
\texttt{Claude 3 Sonnet (icl)} & - & 1.85 (5.29) & 1.25 (3.57) & - & - \\
\texttt{GPT-4 Turbo (icl)} & - & \textbf{5.30} & 1.80 & - & - \\
\textit{ground truth} & - & 5.95 & 6.00 & - & - \\
\bottomrule
\end{tabular}
\begin{tabular}{l|c|cc|cc}
& \textcolor{white}{\textbf{NLL}($\downarrow$)} & \multicolumn{2}{c|}{\textcolor{white}{\textbf{Offline Human Ratings} ($\uparrow$)}} & \multicolumn{2}{c}{\textcolor{white}{\textbf{Online Human Ratings} ($\uparrow$)}} \vspace{-1.5mm}\\
\textbf{Spoken conversations} &  & \textcolor{white}{....}Content\textcolor{white}{....} & \textcolor{white}{.}Timing\textcolor{white}{.} & \textcolor{white}{....} Content\textcolor{white}{....}  & Timing \\
\midrule
\texttt{Pythia 160M (ft)} & 2261 & 0.8 & 1.4 & 0.6 & 0.4 \\
\texttt{Pythia 1.4B (ft)} & 1724 & 2.3 & 3.8 & 1.0 & 1.0 \\
\texttt{Pythia 12B (ft)} & 1661 & 3.1 & 3.8 & 1.6 & 1.8 \\
\texttt{Gemma 2B (ft)} & 1608 & 3.9 & 4.3 & 2.2 & 3.4 \\
\texttt{Llama 2 7B (ft)} & \textbf{1532} & 4.3 & \textbf{4.8} & \textbf{4.0} & \textbf{5.2} \\
\texttt{Claude 3 Sonnet (icl)} & - & 4.2 & 3.7 & - & - \\
\texttt{GPT-4 Turbo (icl)} & - & \textbf{5.0} & 3.8 & - & - \\
\textit{ground truth} & - & 3.7 & 3.9 & - & - \\
\bottomrule
\end{tabular}
\caption{\small \textbf{Instant messenger (\textit{top}) and spoken conversation (\textit{bottom}) quality scores.} \texttt{ft} = finetuned and \texttt{icl} = in-context learning. We compute negative log likelihood per document rather than averaged per token, so that it is comparable across vocabularies. Human ratings range from 0 (worst) to 6 (best). When relevant, we provide scores in parentheses with refusals filtered out. We rate \textit{consistency} (how coherent the conversation is generally) and \textit{fidelity} (how well the model mimics the authors specifically) for instant messenger, and \textit{content} vs. \textit{timing} for speech. See Appendix~\ref{app:more-quantitative} for more details and experiments comparing the ground truth and predicted timestamp distributions.}
\label{tab:joint-quality}
\end{table}

\paragraph*{Overpowering tone} API-based models are tuned to have a particular voice, which bleeds through into the generated messages. So while the conversations are more coherent, they are usually easy to distinguish from the ground truth based on style cues alone. Claude 3 often refuses to perform the task when the chat history discusses politics.

\paragraph*{Speaker consistency}
The finetuned models sometimes struggle to maintain consistent identities for the speakers, mostly across conversations (e.g., one speaker talks about having a sister, when it is only the other speaker who has a sister) but sometimes also within conversations (i.e., a speaker appears to respond to itself). 


\paragraph*{Promise as an evaluation for long context LLMs}
Instant messenger history continuation is a promising task for human evaluation of long in-context learning. Each message history is highly distinct, yet private and therefore guaranteed to be unleaked. While it is prohibitively time-consuming for a human rater to read extremely long prompts in general, if they are instead a participant in the original conversation, they are already deeply familiar with the content and can easily spot errors without additional effort.

\subsection{Spoken Conversations}
\label{subsec:spoken-eval}

As our training dataset, we use a random 1000-hour subset of cases argued before the U.S. Supreme Court, totaling 33,640,559
characters. We sample other cases into a $\sim$350-hour val set and $\sim$295-hour test set. We preprocess the data with WhisperX~\citep{radford2022whisper, bain2022whisperx}, which supports timed diarized word-level ASR. Note that pseudolabeled diarized speech data tends to undercapture timestamp overlap across speakers~\citep{Liesenfeld_2023}, so this data may not reflect fine-grained turn-taking behavior. We lowercase and strip punctuation from the data to make the formatting consistent with streaming ASR.

\subsubsection{Performance}
\label{subsubsec:speech-perf}

See Figure~\ref{fig:joint-performance-reqs} for more details on the performance properties of our spoken conversation control format. 
The highlights are: With $t_{react}$ = 200ms, the 99th percentile is {\speechninetynine} tok/s and 99.9th is {\speechninetyninepointnine} tok/s. On average, the control-formatted token length is 4.3x the plaintext length (median 5x). Note that this ratio is heavily dependent on the way the tokenizer handles digits; many modern tokenizers force individual digits to be separate tokens to improve arithmetic, but in this case, given enough data, 000-999 could reasonably be single tokens. We calculate the rates for this ``optimized tokenizer'': the 99th percentile is {\optimalspeechninetynine} tok/s and 99.9th is {\optimalspeechninetyninepointnine}. On average, the control-formatted token length is 1.8x plaintext length (median 2.0x).

In terms of our proof of concept implementation: we use Google Cloud streaming \href{https://cloud.google.com/speech-to-text?hl=en}{Speech-To-Text} and \href{https://cloud.google.com/text-to-speech?hl=en}{Text-To-Speech} APIs on the client, piped through an \texttt{ssh tty} as text to an A100 40GB server. We measure the end-to-end latency of the former at about 500 ms (from word end to model input) and the latter at about 80 ms; on-device cascade and base models would likely have even lower latency.

\subsubsection{Quality}
\label{subsubsec:speech-quality}

See Table~\ref{tab:joint-quality} for spoken conversation quality results across models; see Appendix~\ref{app:qualitative-examples} for qualitative examples. It is prohibitively time-consuming to read the entire context or each case, and the rater has some legal knowledge but is not an expert, so there may be more of a gap in content quality than is reflected by the scores. In the offline human rating setting, we play the transcripts aloud to judge timing, though with word-level text to speech it is difficult to judge the finer points. Like for instant messanger dialogues, better pretrained models tend to achieve better results. Llama 2 7B (\texttt{ft}) responds remarkably well to turn-taking in the online setting, though there is still obvious room for improvement in all regards.

\section{Related Work}
\label{sec:related-work}

We survey related work in three areas: text dialogues, spoken dialogues, and use of language models to model time broadly.
\subsection{Text Dialogue Modeling}
\label{sec:text-rw}

Modeling text dialogues is perhaps the founding problem of artificial intelligence: Turing's imitation game poses the challenge of distinguishing man from machine through turn-by-turn text dialogue~\citep{turing1950computing}. While timing is mentioned here (a model that responds too quickly could be distinguished from a human), the interaction model is limited. Since then there has been a wealth of work on dialogue systems~\citep{ni2022recent}, initially with complex rule-based methods~\citep{eliza} but shifting over time towards unified deep learning methods, culminating in Meena \& LaMDA~\citep{adiwardana2020humanlike, thoppilan2022lamda}, the Blenderbot series~\citep{roller2020recipes, komeili2021internetaugmented, shuster2022blenderbot}, and of course the recent wave of chatbots such as ChatGPT~\citep{chatgpt}, Gemini~\citep{geminibard}, Copilot~\citep{copilot}, Claude~\citep{claude}, Pi~\citep{pi}, Coral~\citep{coral}, HuggingChat~\citep{huggingchat}, etc. These chatbot works have primarily focused on basic, goal-directed conversational capabilities in the desired domains, which until recently has been very challenging, and less on the interaction model. Replika~\citep{replika} and certain modes in Character.AI~\citep{characterai} do allow multiple messages per conversation turn, but with undisclosed methods and unclear limitations.

CICERO~\citep{cicero} studies Diplomacy, a political strategy game that involves instant messaging with other players in real time. The primary focus is on using dialogue paired with actions to achieve certain goals in the game, which implies the ability to imitate natural timing to avoid raising suspicion with human players. CICERO uses a chain of encoder-decoder models and heuristics to perform tasks such as predicting the next message time vs. content independently, and not all context is available to all models. Messages are rejected/resampled when user input causally intervenes on planned messages. Our work uses a simpler approach with a single transcript in a decoder-only model, which minimizes recomputation and makes all information available for all decisions; we further improve performance by using a reaction time window and causal speculative decoding.

The task of imitating specific people based on their digital footprint (for better or worse) has captured the popular imagination, featuring in shows like \href{https://en.wikipedia.org/wiki/Silicon_Valley_(season_6)#:~:text=%22Artificial%20Lack%20of%20Intelligence%22}{\textit{Silicon Valley}}, \href{https://en.wikipedia.org/wiki/Be_Right_Back}{\textit{Black Mirror}} and \href{https://en.wikipedia.org/wiki/The_Riddle_of_the_Sphinx_(Westworld)}{\textit{Westworld}} and described with names like generative clones or ghosts in academic literature~\citep{morris2024generative}. Blog posts about finetuning LMs on personal chat histories are relatively common, but they either model timed transcipts noninteractively, or synchronous turn by turn conversations interactively (as a traditional chatbot). We are not aware of prior work that turns models of timed transcripts into interactive applications.

\subsection{Spoken Dialogue Modeling}
\label{sec:spoken-rw}

To go beyond manually crafted turn-taking heuristics for what is in generality an extremely complex task~\citep{turntakingsurvey}, the main approach for generating spoken conversations has been direct audio modeling. dGSLM~\citep{nguyen2022generative}, AudioLM~\citep{borsos2023audiolm}, and SpiRit-LM~\citep{nguyen2024spiritlm} do this by modeling learned discrete tokens with autoregressive language models; the former models two streams of audio (dialogues), while the latter two model one. While the token modeling is causal, the tokenization is not, so these methods cannot yet be used in a streaming fashion. 

Discrete audio tokenization is generally performed at a fixed rate of $\sim$40-50 tok/s for a single audio stream, vs. $\sim$20 tok/s for our approach supporting arbitrary numbers of speakers.\footnote{With that said, as with sparse vs. dense approaches generally, under extreme load the bandwidth required for sparse indexing over time may be higher than dense tokenization without indexing. 
And because our approach is sparse over time, it is more difficult to batch and has inconsistent load, which may be disadvantageous for bulk serving.
} This fits into the general pattern of cascaded vs. end-to-end models: cascaded models are generally more performant/require less data and therefore can be developed sooner using fewer resources, but they are eventually superseded by end-to-end models which can provide the optimal quality given sufficient resources.

Though not exactly dialogue, simultaneous translation often operates through a cascade of ASR and TTS, though timing information (besides the relative ordering of words in the source and target streams) is stripped away~\citep{simulspeech, ma2020streaming}.

\subsection{Time-Aware Language Models}
\label{sec:time-lm-rw}

There are many works that make language models aware of time in one sense or another. Even without special effort, language models learn latent representations of time to the extent that it helps explain the training distribution~\citep{gurnee2024language}. The language model CTRL~\citep{keskar2019ctrl} is conditioned on metadata about each document, which may include the publication date. Whisper~\citep{radford2022robust} and some other speech-to-text models predict timestamps as text. \citet{park2023generative} lets loose generative agents in a virtual town environment, where they act on schedules in accordance with the virtual time. Language models have been used as the backbone for time series forecasting, whether pretrained~\citep{das2024decoderonly}, finetuned~\citep{jin2024timellm}, or zero-shot~\citep{gruver2023large}, though here time is usually dense (proceeds at a fixed rate). 
We are not aware of works that model timestamps as text and interpret those timestamps as an input/output stream with respect to the real-world time.

\section{Conclusion}

In this paper, we presented a simple yet general method for simulating real-time interactive conversations using pretrained language models---modeling \textit{timed diarized transcripts} and decoding with \textit{causal rejection sampling}---situated in two use cases: instant messenger dialogues and spoken conversations. It is easy to imagine extensions such as multiple simultaneous conversations with one simulated individual (by adding conversation ids in addition to speaker ids) or modeling multimodal conversations (images, actions, etc.), though this may require more capable language models. While we demonstrated the promise of this method using interactive conversations, it can be applied to turn language models into interactive models for any kind of event sequence, i.e., sparse-over-time world models. We hope that this method will facilitate more flexible interaction with the underlying capabilities of language models and enable new applications in fields such as gaming and entertainment.

\section*{Ethical Considerations}

While work improving the ability to simulate real-time interactive conversations can make language models more useful or delightful, it also poses risks for fraud and manipulation. In order to mitigate these risks, we limit our work to simulating natural conversations in text, a medium which is perceived as less trustworthy than audio or video. (While we simulate the timing aspects of spoken conversation, our generations are still easily distinguished from real speech due to the unnatural sound of word-level text to speech, and custom text-to-speech models would be required to move beyond this.) We provide only proofs of concept with small datasets, and do not scale up to sizes where these capabilities would become more refined. We also do not study goal-directed methods which could be used to steer a model to execute fraud.

We believe that it is valuable to expose this capability overhang so that the community can respond with appropriate measures. For example, a better understanding of the amount of data needed to impersonate someone with a generative clone could affect how much conversational data users are comfortable sharing publicly on social media, or motivate end-to-end encryption/disappearing messages to prevent private data leakage in the event of hacking. Developing interfaces for language models that are not immediately distinguishable from humans could also help to evaluate extreme risks like deception and persuasion in frontier models~\citep{shevlane2023model}, to the extent that people react differently to communication that they perceive to be from a model vs. another person. Bad actors are already capable of sophisticated \href{https://www.cnn.com/2024/02/04/asia/deepfake-cfo-scam-hong-kong-intl-hnk/index.html}{deepfake scams} and aren't exactly forthcoming about their methods.

There are also ethical considerations when simulating real people or fictional characters absent ill intent, such as privacy and the effects of parasocial relationships; these tend to be general concerns that are not strictly related to real-time interactivity. See~\citet{morris2024generative} for an in-depth discussion of these factors. In terms of the specific datasets we used in this paper: We used our own instant messenger history with the consent and active involvement of both participants, and do not release the data/model for privacy reasons.
The U.S. Supreme Court's oral arguments are inherently public and the conversation is in a specialized legal domain rather than anything that would encourage parasocial relationships.
We model only text transcripts and use generic text to speech voices (i.e., we do not contribute methods to impersonate any of the speakers).

\section*{Reproducibility Statement}

We publicly release the code for our case studies at \repolink. We do not release our own personal instant messenger history for reasons of privacy, but you can reproduce the instant messenger case study by bringing your own data. The data for the spoken conversation case study is public and can be reproduced.

\section*{Acknowledgments}

GA is supported by a fellowship from the Kempner Institute for the Study of Natural and Artificial Intelligence at Harvard University.

LMK is supported by the Rhodes Trust.

\bibliography{template}

\begin{thebibliography}{53}
\providecommand{\natexlab}[1]{#1}
\providecommand{\url}[1]{\texttt{#1}}
\expandafter\ifx\csname urlstyle\endcsname\relax
  \providecommand{\doi}[1]{doi: #1}\else
  \providecommand{\doi}{doi: \begingroup \urlstyle{rm}\Url}\fi

\bibitem[Adiwardana et~al.(2020)Adiwardana, Luong, So, Hall, Fiedel, Thoppilan, Yang, Kulshreshtha, Nemade, Lu, and Le]{adiwardana2020humanlike}
Daniel Adiwardana, Minh-Thang Luong, David~R. So, Jamie Hall, Noah Fiedel, Romal Thoppilan, Zi~Yang, Apoorv Kulshreshtha, Gaurav Nemade, Yifeng Lu, and Quoc~V. Le.
\newblock Towards a human-like open-domain chatbot, 2020.

\bibitem[Anthropic()]{claude3}
Anthropic.
\newblock The claude 3 model family: Opus, sonnet, haiku.
\newblock URL \url{https://www-cdn.anthropic.com/de8ba9b01c9ab7cbabf5c33b80b7bbc618857627/Model_Card_Claude_3.pdf}.

\bibitem[Anthropic(2023)]{claude}
Anthropic.
\newblock Introducing {C}laude, 2023.
\newblock URL \url{https://www.anthropic.com/news/introducing-claude}.

\bibitem[Bain et~al.(2023)Bain, Huh, Han, and Zisserman]{bain2022whisperx}
Max Bain, Jaesung Huh, Tengda Han, and Andrew Zisserman.
\newblock Whisperx: Time-accurate speech transcription of long-form audio.
\newblock \emph{INTERSPEECH 2023}, 2023.

\bibitem[Bakhtin et~al.(2022)Bakhtin, Brown, Dinan, Farina, Flaherty, Fried, Goff, Gray, Hu, Jacob, Komeili, Konath, Kwon, Lerer, Lewis, Miller, Mitts, Renduchintala, Roller, Rowe, Shi, Spisak, Wei, Wu, Zhang, and Zijlstra]{cicero}
Anton Bakhtin, Noam Brown, Emily Dinan, Gabriele Farina, Colin Flaherty, Daniel Fried, Andrew Goff, Jonathan Gray, Hengyuan Hu, Athul~Paul Jacob, Mojtaba Komeili, Karthik Konath, Minae Kwon, Adam Lerer, Mike Lewis, Alexander~H. Miller, Sasha Mitts, Adithya Renduchintala, Stephen Roller, Dirk Rowe, Weiyan Shi, Joe Spisak, Alexander Wei, David Wu, Hugh Zhang, and Markus Zijlstra.
\newblock Human-level play in the game of <i>diplomacy</i> by combining language models with strategic reasoning.
\newblock \emph{Science}, 378\penalty0 (6624):\penalty0 1067--1074, 2022.
\newblock \doi{10.1126/science.ade9097}.
\newblock URL \url{https://www.science.org/doi/abs/10.1126/science.ade9097}.

\bibitem[Biderman et~al.(2023)Biderman, Schoelkopf, Anthony, Bradley, O'Brien, Hallahan, Khan, Purohit, Prashanth, Raff, Skowron, Sutawika, and van~der Wal]{biderman2023pythia}
Stella Biderman, Hailey Schoelkopf, Quentin Anthony, Herbie Bradley, Kyle O'Brien, Eric Hallahan, Mohammad~Aflah Khan, Shivanshu Purohit, USVSN~Sai Prashanth, Edward Raff, Aviya Skowron, Lintang Sutawika, and Oskar van~der Wal.
\newblock Pythia: A suite for analyzing large language models across training and scaling, 2023.
\newblock URL \url{https://arxiv.org/abs/2304.01373}.

\bibitem[Borsos et~al.(2023)Borsos, Marinier, Vincent, Kharitonov, Pietquin, Sharifi, Roblek, Teboul, Grangier, Tagliasacchi, and Zeghidour]{borsos2023audiolm}
Zalán Borsos, Raphaël Marinier, Damien Vincent, Eugene Kharitonov, Olivier Pietquin, Matt Sharifi, Dominik Roblek, Olivier Teboul, David Grangier, Marco Tagliasacchi, and Neil Zeghidour.
\newblock Audiolm: a language modeling approach to audio generation, 2023.

\bibitem[Boyle(2019)]{scotusrepo}
Walker Boyle.
\newblock Us supreme court annotated transcripts (auto-updated), 2019.
\newblock URL \url{https://github.com/walkerdb/supreme_court_transcripts}.

\bibitem[character.ai()]{characterai}
character.ai.
\newblock New feature announcement: Character group chat.
\newblock URL \url{https://blog.character.ai/new-feature-announcement-character-group-chat/}.

\bibitem[Chen et~al.(2023)Chen, Borgeaud, Irving, Lespiau, Sifre, and Jumper]{chen2023accelerating}
Charlie Chen, Sebastian Borgeaud, Geoffrey Irving, Jean-Baptiste Lespiau, Laurent Sifre, and John Jumper.
\newblock Accelerating large language model decoding with speculative sampling, 2023.

\bibitem[Cohere()]{coral}
Cohere.
\newblock Introducing coral, the knowledge assistant for enterprises.
\newblock URL \url{https://txt.cohere.com/introducing-coral/}.

\bibitem[Das et~al.(2024)Das, Kong, Sen, and Zhou]{das2024decoderonly}
Abhimanyu Das, Weihao Kong, Rajat Sen, and Yichen Zhou.
\newblock A decoder-only foundation model for time-series forecasting, 2024.

\bibitem[Google(2024)]{geminibard}
Google.
\newblock Bard becomes gemini: Try ultra 1.0 and a new mobile app today, 2024.
\newblock URL \url{https://blog.google/products/gemini/bard-gemini-advanced-app/}.

\bibitem[Gruver et~al.(2023)Gruver, Finzi, Qiu, and Wilson]{gruver2023large}
Nate Gruver, Marc Finzi, Shikai Qiu, and Andrew~Gordon Wilson.
\newblock Large language models are zero-shot time series forecasters, 2023.

\bibitem[Gurnee \& Tegmark(2024)Gurnee and Tegmark]{gurnee2024language}
Wes Gurnee and Max Tegmark.
\newblock Language models represent space and time, 2024.

\bibitem[Ho \& Salimans(2022)Ho and Salimans]{ho2022classifierfree}
Jonathan Ho and Tim Salimans.
\newblock Classifier-free diffusion guidance, 2022.

\bibitem[Holtzman et~al.(2020)Holtzman, Buys, Du, Forbes, and Choi]{nucleus_sampling}
Ari Holtzman, Jan Buys, Li~Du, Maxwell Forbes, and Yejin Choi.
\newblock The curious case of neural text degeneration, 2020.

\bibitem[HuggingFace()]{huggingchat}
HuggingFace.
\newblock Huggingchat.
\newblock URL \url{https://huggingface.co/chat/privacy}.

\bibitem[Inflection()]{pi}
Inflection.
\newblock Introducing pi, your personal ai.
\newblock URL \url{https://inflection.ai/press}.

\bibitem[Jain et~al.()Jain, Bansal, Avnish, and Singh]{humanAudioReactTime}
Aditya Jain, Ramta Bansal, Avnish, and KD~Singh.
\newblock A comparative study of visual and auditory reaction times on the basis of gender and physical activity levels of medical first year students.
\newblock \doi{10.4103/2229-516X.157168}.
\newblock URL \url{https://www.ncbi.nlm.nih.gov/pmc/articles/PMC4456887/}.

\bibitem[Jin et~al.(2024)Jin, Wang, Ma, Chu, Zhang, Shi, Chen, Liang, Li, Pan, and Wen]{jin2024timellm}
Ming Jin, Shiyu Wang, Lintao Ma, Zhixuan Chu, James~Y. Zhang, Xiaoming Shi, Pin-Yu Chen, Yuxuan Liang, Yuan-Fang Li, Shirui Pan, and Qingsong Wen.
\newblock Time-llm: Time series forecasting by reprogramming large language models, 2024.

\bibitem[Keskar et~al.(2019)Keskar, McCann, Varshney, Xiong, and Socher]{keskar2019ctrl}
Nitish~Shirish Keskar, Bryan McCann, Lav~R. Varshney, Caiming Xiong, and Richard Socher.
\newblock Ctrl: A conditional transformer language model for controllable generation, 2019.

\bibitem[Komeili et~al.(2021)Komeili, Shuster, and Weston]{komeili2021internetaugmented}
Mojtaba Komeili, Kurt Shuster, and Jason Weston.
\newblock Internet-augmented dialogue generation, 2021.

\bibitem[Leviathan et~al.(2023)Leviathan, Kalman, and Matias]{leviathan2023fast}
Yaniv Leviathan, Matan Kalman, and Yossi Matias.
\newblock Fast inference from transformers via speculative decoding, 2023.

\bibitem[Liesenfeld et~al.(2023)Liesenfeld, Lopez, and Dingemanse]{Liesenfeld_2023}
Andreas Liesenfeld, Alianda Lopez, and Mark Dingemanse.
\newblock The timing bottleneck: Why timing and overlap are mission-critical for conversational user interfaces, speech recognition and dialogue systems.
\newblock In \emph{Proceedings of the 24th Meeting of the Special Interest Group on Discourse and Dialogue}. Association for Computational Linguistics, 2023.
\newblock \doi{10.18653/v1/2023.sigdial-1.45}.
\newblock URL \url{http://dx.doi.org/10.18653/v1/2023.sigdial-1.45}.

\bibitem[Ma et~al.(2020{\natexlab{a}})Ma, Zheng, Liu, Zheng, Liu, Peng, Church, and Huang]{ma2020incremental}
Mingbo Ma, Baigong Zheng, Kaibo Liu, Renjie Zheng, Hairong Liu, Kainan Peng, Kenneth Church, and Liang Huang.
\newblock Incremental text-to-speech synthesis with prefix-to-prefix framework, 2020{\natexlab{a}}.

\bibitem[Ma et~al.(2020{\natexlab{b}})Ma, Wang, Dousti, Koehn, and Pino]{ma2020streaming}
Xutai Ma, Yongqiang Wang, Mohammad~Javad Dousti, Philipp Koehn, and Juan Pino.
\newblock Streaming simultaneous speech translation with augmented memory transformer, 2020{\natexlab{b}}.

\bibitem[Microsoft()]{copilot}
Microsoft.
\newblock Announcing microsoft copilot, your everyday ai companion.
\newblock URL \url{https://blogs.microsoft.com/blog/2023/09/21/announcing-microsoft-copilot-your-everyday-ai-companion/}.

\bibitem[Morris \& Brubaker(2024)Morris and Brubaker]{morris2024generative}
Meredith~Ringel Morris and Jed~R. Brubaker.
\newblock Generative ghosts: Anticipating benefits and risks of ai afterlives, 2024.
\newblock URL \url{https://arxiv.org/abs/2402.01662}.

\bibitem[Nguyen et~al.(2022)Nguyen, Kharitonov, Copet, Adi, Hsu, Elkahky, Tomasello, Algayres, Sagot, Mohamed, and Dupoux]{nguyen2022generative}
Tu~Anh Nguyen, Eugene Kharitonov, Jade Copet, Yossi Adi, Wei-Ning Hsu, Ali Elkahky, Paden Tomasello, Robin Algayres, Benoit Sagot, Abdelrahman Mohamed, and Emmanuel Dupoux.
\newblock Generative spoken dialogue language modeling, 2022.
\newblock URL \url{https://arxiv.org/abs/2203.16502}.

\bibitem[Nguyen et~al.(2024)Nguyen, Muller, Yu, Costa-jussa, Elbayad, Popuri, Duquenne, Algayres, Mavlyutov, Gat, Synnaeve, Pino, Sagot, and Dupoux]{nguyen2024spiritlm}
Tu~Anh Nguyen, Benjamin Muller, Bokai Yu, Marta~R. Costa-jussa, Maha Elbayad, Sravya Popuri, Paul-Ambroise Duquenne, Robin Algayres, Ruslan Mavlyutov, Itai Gat, Gabriel Synnaeve, Juan Pino, Benoit Sagot, and Emmanuel Dupoux.
\newblock Spirit-lm: Interleaved spoken and written language model, 2024.

\bibitem[Ni et~al.(2022)Ni, Young, Pandelea, Xue, and Cambria]{ni2022recent}
Jinjie Ni, Tom Young, Vlad Pandelea, Fuzhao Xue, and Erik Cambria.
\newblock Recent advances in deep learning based dialogue systems: A systematic survey, 2022.

\bibitem[OpenAI()]{gpt4turbo}
OpenAI.
\newblock New models and developer products announced at {D}ev{D}ay.
\newblock URL \url{https://openai.com/blog/new-models-and-developer-products-announced-at-devday}.

\bibitem[OpenAI(2023)]{chatgptaudio}
OpenAI.
\newblock Chatgpt can now see, hear, and speak, 2023.
\newblock URL \url{https://openai.com/blog/chatgpt-can-now-see-hear-and-speak}.

\bibitem[Park et~al.(2023)Park, O'Brien, Cai, Morris, Liang, and Bernstein]{park2023generative}
Joon~Sung Park, Joseph~C. O'Brien, Carrie~J. Cai, Meredith~Ringel Morris, Percy Liang, and Michael~S. Bernstein.
\newblock Generative agents: Interactive simulacra of human behavior, 2023.
\newblock URL \url{https://arxiv.org/abs/2304.03442}.

\bibitem[Radford et~al.(2022{\natexlab{a}})Radford, Kim, Xu, Brockman, McLeavey, and Sutskever]{radford2022robust}
Alec Radford, Jong~Wook Kim, Tao Xu, Greg Brockman, Christine McLeavey, and Ilya Sutskever.
\newblock Robust speech recognition via large-scale weak supervision, 2022{\natexlab{a}}.

\bibitem[Radford et~al.(2022{\natexlab{b}})Radford, Kim, Xu, Brockman, McLeavey, and Sutskever]{radford2022whisper}
Alec Radford, Jong~Wook Kim, Tao Xu, Greg Brockman, Christine McLeavey, and Ilya Sutskever.
\newblock Robust speech recognition via large-scale weak supervision, 2022{\natexlab{b}}.
\newblock URL \url{https://arxiv.org/abs/2212.04356}.

\bibitem[Reid et~al.(2024)Reid, Savinov, Teplyashin, Lepikhin, Lillicrap, baptiste Alayrac, Soricut, Lazaridou, Firat, Schrittwieser, Antonoglou, Anil, Borgeaud, Dai, Millican, Dyer, Glaese, Sottiaux, Lee, Viola, Reynolds, Xu, Molloy, Chen, Isard, Barham, Hennigan, McIlroy, Johnson, Schalkwyk, Collins, Rutherford, Moreira, Ayoub, Goel, Meyer, Thornton, Yang, Michalewski, Abbas, Schucher, Anand, Ives, Keeling, Lenc, Haykal, Shakeri, Shyam, Chowdhery, Ring, Spencer, Sezener, Vilnis, Chang, Morioka, Tucker, Zheng, Woodman, Attaluri, Kocisky, Eltyshev, Chen, Chung, Selo, Brahma, Georgiev, Slone, Zhu, Lottes, Qiao, Caine, Riedel, Tomala, Chadwick, Love, Choy, Mittal, Houlsby, Tang, Lamm, Bai, Zhang, He, Cheng, Humphreys, Li, Brin, Cassirer, Miao, Zilka, Tobin, Xu, Proleev, Sohn, Magni, Hendricks, Gao, Ontañón, Bunyan, Byrd, Sharma, Zhang, Pinto, Sinha, Mehta, Jia, Caelles, Webson, Morris, Roelofs, Ding, Strudel, Xiong, Ritter, Dehghani, Chaabouni, Karmarkar, Lai, Mentzer, Xu, Li, Zhang, Paine, Goldin, Neyshabur,
  Baumli, Levskaya, Laskin, Jia, Rae, Xiao, He, Giordano, Yagati, Lespiau, Natsev, Ganapathy, Liu, Martins, Chen, Xu, Barnes, May, Vezer, Oh, Franko, Bridgers, Zhao, Wu, Mustafa, Sechrist, Parisotto, Pillai, Larkin, Gu, Sorokin, Krikun, Guseynov, Landon, Datta, Pritzel, Thacker, Yang, Hui, Hauth, Yeh, Barker, Mao-Jones, Austin, Sheahan, Schuh, Svensson, Jain, Ramasesh, Briukhov, Chung, von Glehn, Butterfield, Jhakra, Wiethoff, Frye, Grimstad, Changpinyo, Lan, Bortsova, Wu, Voigtlaender, Sainath, Smith, Hawkins, Cao, Besley, Srinivasan, Omernick, Gaffney, Surita, Burnell, Damoc, Ahn, Brock, Pajarskas, Petrushkina, Noury, Blanco, Swersky, Ahuja, Avrahami, Misra, de~Liedekerke, Iinuma, Polozov, York, van~den Driessche, Michel, Chiu, Blevins, Gleicher, Recasens, Rrustemi, Gribovskaya, Roy, Gworek, Arnold, Lee, Lee-Thorp, Maggioni, Piqueras, Badola, Vikram, Gonzalez, Baddepudi, Senter, Devlin, Qin, Azzam, Trebacz, Polacek, Krishnakumar, yiin Chang, Tung, Penchev, Joshi, Olszewska, Muir, Wirth, Hartman, Newlan,
  Kashem, Bolina, Dabir, van Amersfoort, Ahmed, Cobon-Kerr, Kamath, Hrafnkelsson, Hou, Mackinnon, Frechette, Noland, Si, Taropa, Li, Crone, Gulati, Cevey, Adler, Ma, Silver, Tokumine, Powell, Lee, Chang, Hassan, Mincu, Yang, Levine, Brennan, Wang, Hodkinson, Zhao, Lipschultz, Pope, Chang, Li, Shafey, Paganini, Douglas, Bohnet, Pardo, Odoom, Rosca, dos Santos, Soparkar, Guez, Hudson, Hansen, Asawaroengchai, Addanki, Yu, Stokowiec, Khan, Gilmer, Lee, Bostock, Rong, Caton, Pejman, Pavetic, Brown, Sharma, Lučić, Samuel, Djolonga, Mandhane, Sjösund, Buchatskaya, White, Clay, Jiang, Lim, Hemsley, Labanowski, Cao, Steiner, Hashemi, Austin, Gergely, Blyth, Stanton, Shivakumar, Siddhant, Andreassen, Araya, Sethi, Shivanna, Hand, Bapna, Khodaei, Miech, Tanzer, Swing, Thakoor, Pan, Nado, Winkler, Yu, Saleh, Maggiore, Barr, Giang, Kagohara, Danihelka, Marathe, Feinberg, Elhawaty, Ghelani, Horgan, Miller, Walker, Tanburn, Tariq, Shrivastava, Xia, Chiu, Ashwood, Baatarsukh, Samangooei, Alcober, Stjerngren, Komarek,
  Tsihlas, Boral, Comanescu, Chen, Liu, Bloxwich, Chen, Sun, Feng, Mauger, Dotiwalla, Hellendoorn, Sharman, Zheng, Haridasan, Barth-Maron, Swanson, Rogozińska, Andreev, Rubenstein, Sang, Hurt, Elsayed, Wang, Lacey, Ilić, Zhao, Aroyo, Iwuanyanwu, Nikolaev, Lakshminarayanan, Jazayeri, Kaufman, Varadarajan, Tekur, Fritz, Khalman, Reitter, Dasgupta, Sarcar, Ornduff, Snaider, Huot, Jia, Kemp, Trdin, Vijayakumar, Kim, Angermueller, Lao, Liu, Zhang, Engel, Greene, White, Austin, Taylor, Ashraf, Liu, Georgaki, Cai, Kulizhskaya, Goenka, Saeta, Vodrahalli, Frank, de~Cesare, Robenek, Richardson, Alnahlawi, Yew, Ponnapalli, Tagliasacchi, Korchemniy, Kim, Li, Rosgen, Ashwood, Levin, Wiesner, Banzal, Srinivasan, Yu, Çağlar Ünlü, Reid, Tung, Finchelstein, Kumar, Elisseeff, Huang, Zhang, Zhu, Aguilar, Giménez, Xia, Dousse, Gierke, Yeganeh, Yates, Jalan, Li, Latorre-Chimoto, Nguyen, Durden, Kallakuri, Liu, Johnson, Tsai, Talbert, Liu, Neitz, Elkind, Selvi, Jasarevic, Soares, Cui, Wang, Wang, Ye, Kallarackal, Loher,
  Lam, Broder, Holtmann-Rice, Martin, Ramadhana, Toyama, Shukla, Basu, Mohan, Fernando, Fiedel, Paterson, Li, Garg, Park, Choi, Wu, Singh, Zhang, Globerson, Yu, Carpenter, de~Chaumont~Quitry, Radebaugh, Lin, Tudor, Shroff, Garmon, Du, Vats, Lu, Iqbal, Yakubovich, Tripuraneni, Manyika, Qureshi, Hua, Ngani, Raad, Forbes, Bulanova, Stanway, Sundararajan, Ungureanu, Bishop, Li, Venkatraman, Li, Thornton, Scellato, Gupta, Wang, Tenney, Wu, Shenoy, Carvajal, Wright, Bariach, Xiao, Hawkins, Dalmia, Farabet, Valenzuela, Yuan, Welty, Agarwal, Chen, Kim, Hulse, Dukkipati, Paszke, Bolt, Davoodi, Choo, Beattie, Prendki, Vashisht, Santamaria-Fernandez, Cobo, Wilkiewicz, Madras, Elqursh, Uy, Ramirez, Harvey, Liechty, Zen, Seibert, Hu, Elhawaty, Khorlin, Le, Aharoni, Li, Wang, Kumar, Lince, Casagrande, Hoover, Badawy, Soergel, Vnukov, Miecnikowski, Simsa, Koop, Kumar, Sellam, Vlasic, Daruki, Shabat, Zhang, Su, Zhang, Liu, Sun, Palmer, Ghaffarkhah, Xiong, Cotruta, Fink, Dixon, Sreevatsa, Goedeckemeyer, Dimitriev, Jafari,
  Crocker, FitzGerald, Kumar, Ghemawat, Philips, Liu, Liang, Sterneck, Repina, Wu, Knight, Georgiev, Lee, Askham, Chakladar, Louis, Crous, Cate, Petrova, Quinn, Owusu-Afriyie, Singhal, Wei, Kim, Vincent, Nasr, Choquette-Choo, Tojo, Lu, de~Las~Casas, Cheng, Bolukbasi, Lee, Fatehi, Ananthanarayanan, Patel, Kaed, Li, Sygnowski, Belle, Chen, Konzelmann, Põder, Garg, Koverkathu, Brown, Dyer, Liu, Nova, Xu, Petrov, Hassabis, Kavukcuoglu, Dean, and Vinyals]{gemini1p5}
Machel Reid, Nikolay Savinov, Denis Teplyashin, Dmitry Lepikhin, Timothy Lillicrap, Jean baptiste Alayrac, Radu Soricut, Angeliki Lazaridou, Orhan Firat, Julian Schrittwieser, Ioannis Antonoglou, Rohan Anil, Sebastian Borgeaud, Andrew Dai, Katie Millican, Ethan Dyer, Mia Glaese, Thibault Sottiaux, Benjamin Lee, Fabio Viola, Malcolm Reynolds, Yuanzhong Xu, James Molloy, Jilin Chen, Michael Isard, Paul Barham, Tom Hennigan, Ross McIlroy, Melvin Johnson, Johan Schalkwyk, Eli Collins, Eliza Rutherford, Erica Moreira, Kareem Ayoub, Megha Goel, Clemens Meyer, Gregory Thornton, Zhen Yang, Henryk Michalewski, Zaheer Abbas, Nathan Schucher, Ankesh Anand, Richard Ives, James Keeling, Karel Lenc, Salem Haykal, Siamak Shakeri, Pranav Shyam, Aakanksha Chowdhery, Roman Ring, Stephen Spencer, Eren Sezener, Luke Vilnis, Oscar Chang, Nobuyuki Morioka, George Tucker, Ce~Zheng, Oliver Woodman, Nithya Attaluri, Tomas Kocisky, Evgenii Eltyshev, Xi~Chen, Timothy Chung, Vittorio Selo, Siddhartha Brahma, Petko Georgiev, Ambrose
  Slone, Zhenkai Zhu, James Lottes, Siyuan Qiao, Ben Caine, Sebastian Riedel, Alex Tomala, Martin Chadwick, Juliette Love, Peter Choy, Sid Mittal, Neil Houlsby, Yunhao Tang, Matthew Lamm, Libin Bai, Qiao Zhang, Luheng He, Yong Cheng, Peter Humphreys, Yujia Li, Sergey Brin, Albin Cassirer, Yingjie Miao, Lukas Zilka, Taylor Tobin, Kelvin Xu, Lev Proleev, Daniel Sohn, Alberto Magni, Lisa~Anne Hendricks, Isabel Gao, Santiago Ontañón, Oskar Bunyan, Nathan Byrd, Abhanshu Sharma, Biao Zhang, Mario Pinto, Rishika Sinha, Harsh Mehta, Dawei Jia, Sergi Caelles, Albert Webson, Alex Morris, Becca Roelofs, Yifan Ding, Robin Strudel, Xuehan Xiong, Marvin Ritter, Mostafa Dehghani, Rahma Chaabouni, Abhijit Karmarkar, Guangda Lai, Fabian Mentzer, Bibo Xu, YaGuang Li, Yujing Zhang, Tom~Le Paine, Alex Goldin, Behnam Neyshabur, Kate Baumli, Anselm Levskaya, Michael Laskin, Wenhao Jia, Jack~W. Rae, Kefan Xiao, Antoine He, Skye Giordano, Lakshman Yagati, Jean-Baptiste Lespiau, Paul Natsev, Sanjay Ganapathy, Fangyu Liu, Danilo
  Martins, Nanxin Chen, Yunhan Xu, Megan Barnes, Rhys May, Arpi Vezer, Junhyuk Oh, Ken Franko, Sophie Bridgers, Ruizhe Zhao, Boxi Wu, Basil Mustafa, Sean Sechrist, Emilio Parisotto, Thanumalayan~Sankaranarayana Pillai, Chris Larkin, Chenjie Gu, Christina Sorokin, Maxim Krikun, Alexey Guseynov, Jessica Landon, Romina Datta, Alexander Pritzel, Phoebe Thacker, Fan Yang, Kevin Hui, Anja Hauth, Chih-Kuan Yeh, David Barker, Justin Mao-Jones, Sophia Austin, Hannah Sheahan, Parker Schuh, James Svensson, Rohan Jain, Vinay Ramasesh, Anton Briukhov, Da-Woon Chung, Tamara von Glehn, Christina Butterfield, Priya Jhakra, Matthew Wiethoff, Justin Frye, Jordan Grimstad, Beer Changpinyo, Charline~Le Lan, Anna Bortsova, Yonghui Wu, Paul Voigtlaender, Tara Sainath, Charlotte Smith, Will Hawkins, Kris Cao, James Besley, Srivatsan Srinivasan, Mark Omernick, Colin Gaffney, Gabriela Surita, Ryan Burnell, Bogdan Damoc, Junwhan Ahn, Andrew Brock, Mantas Pajarskas, Anastasia Petrushkina, Seb Noury, Lorenzo Blanco, Kevin Swersky, Arun
  Ahuja, Thi Avrahami, Vedant Misra, Raoul de~Liedekerke, Mariko Iinuma, Alex Polozov, Sarah York, George van~den Driessche, Paul Michel, Justin Chiu, Rory Blevins, Zach Gleicher, Adrià Recasens, Alban Rrustemi, Elena Gribovskaya, Aurko Roy, Wiktor Gworek, Séb Arnold, Lisa Lee, James Lee-Thorp, Marcello Maggioni, Enrique Piqueras, Kartikeya Badola, Sharad Vikram, Lucas Gonzalez, Anirudh Baddepudi, Evan Senter, Jacob Devlin, James Qin, Michael Azzam, Maja Trebacz, Martin Polacek, Kashyap Krishnakumar, Shuo yiin Chang, Matthew Tung, Ivo Penchev, Rishabh Joshi, Kate Olszewska, Carrie Muir, Mateo Wirth, Ale~Jakse Hartman, Josh Newlan, Sheleem Kashem, Vijay Bolina, Elahe Dabir, Joost van Amersfoort, Zafarali Ahmed, James Cobon-Kerr, Aishwarya Kamath, Arnar~Mar Hrafnkelsson, Le~Hou, Ian Mackinnon, Alexandre Frechette, Eric Noland, Xiance Si, Emanuel Taropa, Dong Li, Phil Crone, Anmol Gulati, Sébastien Cevey, Jonas Adler, Ada Ma, David Silver, Simon Tokumine, Richard Powell, Stephan Lee, Michael Chang, Samer
  Hassan, Diana Mincu, Antoine Yang, Nir Levine, Jenny Brennan, Mingqiu Wang, Sarah Hodkinson, Jeffrey Zhao, Josh Lipschultz, Aedan Pope, Michael~B. Chang, Cheng Li, Laurent~El Shafey, Michela Paganini, Sholto Douglas, Bernd Bohnet, Fabio Pardo, Seth Odoom, Mihaela Rosca, Cicero~Nogueira dos Santos, Kedar Soparkar, Arthur Guez, Tom Hudson, Steven Hansen, Chulayuth Asawaroengchai, Ravi Addanki, Tianhe Yu, Wojciech Stokowiec, Mina Khan, Justin Gilmer, Jaehoon Lee, Carrie~Grimes Bostock, Keran Rong, Jonathan Caton, Pedram Pejman, Filip Pavetic, Geoff Brown, Vivek Sharma, Mario Lučić, Rajkumar Samuel, Josip Djolonga, Amol Mandhane, Lars~Lowe Sjösund, Elena Buchatskaya, Elspeth White, Natalie Clay, Jiepu Jiang, Hyeontaek Lim, Ross Hemsley, Jane Labanowski, Nicola~De Cao, David Steiner, Sayed~Hadi Hashemi, Jacob Austin, Anita Gergely, Tim Blyth, Joe Stanton, Kaushik Shivakumar, Aditya Siddhant, Anders Andreassen, Carlos Araya, Nikhil Sethi, Rakesh Shivanna, Steven Hand, Ankur Bapna, Ali Khodaei, Antoine Miech,
  Garrett Tanzer, Andy Swing, Shantanu Thakoor, Zhufeng Pan, Zachary Nado, Stephanie Winkler, Dian Yu, Mohammad Saleh, Loren Maggiore, Iain Barr, Minh Giang, Thais Kagohara, Ivo Danihelka, Amit Marathe, Vladimir Feinberg, Mohamed Elhawaty, Nimesh Ghelani, Dan Horgan, Helen Miller, Lexi Walker, Richard Tanburn, Mukarram Tariq, Disha Shrivastava, Fei Xia, Chung-Cheng Chiu, Zoe Ashwood, Khuslen Baatarsukh, Sina Samangooei, Fred Alcober, Axel Stjerngren, Paul Komarek, Katerina Tsihlas, Anudhyan Boral, Ramona Comanescu, Jeremy Chen, Ruibo Liu, Dawn Bloxwich, Charlie Chen, Yanhua Sun, Fangxiaoyu Feng, Matthew Mauger, Xerxes Dotiwalla, Vincent Hellendoorn, Michael Sharman, Ivy Zheng, Krishna Haridasan, Gabe Barth-Maron, Craig Swanson, Dominika Rogozińska, Alek Andreev, Paul~Kishan Rubenstein, Ruoxin Sang, Dan Hurt, Gamaleldin Elsayed, Renshen Wang, Dave Lacey, Anastasija Ilić, Yao Zhao, Lora Aroyo, Chimezie Iwuanyanwu, Vitaly Nikolaev, Balaji Lakshminarayanan, Sadegh Jazayeri, Raphaël~Lopez Kaufman, Mani
  Varadarajan, Chetan Tekur, Doug Fritz, Misha Khalman, David Reitter, Kingshuk Dasgupta, Shourya Sarcar, Tina Ornduff, Javier Snaider, Fantine Huot, Johnson Jia, Rupert Kemp, Nejc Trdin, Anitha Vijayakumar, Lucy Kim, Christof Angermueller, Li~Lao, Tianqi Liu, Haibin Zhang, David Engel, Somer Greene, Anaïs White, Jessica Austin, Lilly Taylor, Shereen Ashraf, Dangyi Liu, Maria Georgaki, Irene Cai, Yana Kulizhskaya, Sonam Goenka, Brennan Saeta, Kiran Vodrahalli, Christian Frank, Dario de~Cesare, Brona Robenek, Harry Richardson, Mahmoud Alnahlawi, Christopher Yew, Priya Ponnapalli, Marco Tagliasacchi, Alex Korchemniy, Yelin Kim, Dinghua Li, Bill Rosgen, Zoe Ashwood, Kyle Levin, Jeremy Wiesner, Praseem Banzal, Praveen Srinivasan, Hongkun Yu, Çağlar Ünlü, David Reid, Zora Tung, Daniel Finchelstein, Ravin Kumar, Andre Elisseeff, Jin Huang, Ming Zhang, Rui Zhu, Ricardo Aguilar, Mai Giménez, Jiawei Xia, Olivier Dousse, Willi Gierke, Soheil~Hassas Yeganeh, Damion Yates, Komal Jalan, Lu~Li, Eri Latorre-Chimoto,
  Duc~Dung Nguyen, Ken Durden, Praveen Kallakuri, Yaxin Liu, Matthew Johnson, Tomy Tsai, Alice Talbert, Jasmine Liu, Alexander Neitz, Chen Elkind, Marco Selvi, Mimi Jasarevic, Livio~Baldini Soares, Albert Cui, Pidong Wang, Alek~Wenjiao Wang, Xinyu Ye, Krystal Kallarackal, Lucia Loher, Hoi Lam, Josef Broder, Dan Holtmann-Rice, Nina Martin, Bramandia Ramadhana, Daniel Toyama, Mrinal Shukla, Sujoy Basu, Abhi Mohan, Nick Fernando, Noah Fiedel, Kim Paterson, Hui Li, Ankush Garg, Jane Park, DongHyun Choi, Diane Wu, Sankalp Singh, Zhishuai Zhang, Amir Globerson, Lily Yu, John Carpenter, Félix de~Chaumont~Quitry, Carey Radebaugh, Chu-Cheng Lin, Alex Tudor, Prakash Shroff, Drew Garmon, Dayou Du, Neera Vats, Han Lu, Shariq Iqbal, Alex Yakubovich, Nilesh Tripuraneni, James Manyika, Haroon Qureshi, Nan Hua, Christel Ngani, Maria~Abi Raad, Hannah Forbes, Anna Bulanova, Jeff Stanway, Mukund Sundararajan, Victor Ungureanu, Colton Bishop, Yunjie Li, Balaji Venkatraman, Bo~Li, Chloe Thornton, Salvatore Scellato, Nishesh
  Gupta, Yicheng Wang, Ian Tenney, Xihui Wu, Ashish Shenoy, Gabriel Carvajal, Diana~Gage Wright, Ben Bariach, Zhuyun Xiao, Peter Hawkins, Sid Dalmia, Clement Farabet, Pedro Valenzuela, Quan Yuan, Chris Welty, Ananth Agarwal, Mia Chen, Wooyeol Kim, Brice Hulse, Nandita Dukkipati, Adam Paszke, Andrew Bolt, Elnaz Davoodi, Kiam Choo, Jennifer Beattie, Jennifer Prendki, Harsha Vashisht, Rebeca Santamaria-Fernandez, Luis~C. Cobo, Jarek Wilkiewicz, David Madras, Ali Elqursh, Grant Uy, Kevin Ramirez, Matt Harvey, Tyler Liechty, Heiga Zen, Jeff Seibert, Clara~Huiyi Hu, Mohamed Elhawaty, Andrey Khorlin, Maigo Le, Asaf Aharoni, Megan Li, Lily Wang, Sandeep Kumar, Alejandro Lince, Norman Casagrande, Jay Hoover, Dalia~El Badawy, David Soergel, Denis Vnukov, Matt Miecnikowski, Jiri Simsa, Anna Koop, Praveen Kumar, Thibault Sellam, Daniel Vlasic, Samira Daruki, Nir Shabat, John Zhang, Guolong Su, Jiageng Zhang, Jeremiah Liu, Yi~Sun, Evan Palmer, Alireza Ghaffarkhah, Xi~Xiong, Victor Cotruta, Michael Fink, Lucas Dixon,
  Ashwin Sreevatsa, Adrian Goedeckemeyer, Alek Dimitriev, Mohsen Jafari, Remi Crocker, Nicholas FitzGerald, Aviral Kumar, Sanjay Ghemawat, Ivan Philips, Frederick Liu, Yannie Liang, Rachel Sterneck, Alena Repina, Marcus Wu, Laura Knight, Marin Georgiev, Hyo Lee, Harry Askham, Abhishek Chakladar, Annie Louis, Carl Crous, Hardie Cate, Dessie Petrova, Michael Quinn, Denese Owusu-Afriyie, Achintya Singhal, Nan Wei, Solomon Kim, Damien Vincent, Milad Nasr, Christopher~A. Choquette-Choo, Reiko Tojo, Shawn Lu, Diego de~Las~Casas, Yuchung Cheng, Tolga Bolukbasi, Katherine Lee, Saaber Fatehi, Rajagopal Ananthanarayanan, Miteyan Patel, Charbel Kaed, Jing Li, Jakub Sygnowski, Shreyas~Rammohan Belle, Zhe Chen, Jaclyn Konzelmann, Siim Põder, Roopal Garg, Vinod Koverkathu, Adam Brown, Chris Dyer, Rosanne Liu, Azade Nova, Jun Xu, Slav Petrov, Demis Hassabis, Koray Kavukcuoglu, Jeffrey Dean, and Oriol Vinyals.
\newblock Gemini 1.5: Unlocking multimodal understanding across millions of tokens of context, 2024.
\newblock URL \url{https://arxiv.org/abs/2403.05530}.

\bibitem[Ren et~al.(2020)Ren, Liu, Tan, Zhang, Qin, Zhao, and Liu]{simulspeech}
Yi~Ren, Jinglin Liu, Xu~Tan, Chen Zhang, Tao Qin, Zhou Zhao, and Tie-Yan Liu.
\newblock {S}imul{S}peech: End-to-end simultaneous speech to text translation.
\newblock In Dan Jurafsky, Joyce Chai, Natalie Schluter, and Joel Tetreault (eds.), \emph{Proceedings of the 58th Annual Meeting of the Association for Computational Linguistics}, pp.\  3787--3796, Online, July 2020. Association for Computational Linguistics.
\newblock \doi{10.18653/v1/2020.acl-main.350}.
\newblock URL \url{https://aclanthology.org/2020.acl-main.350}.

\bibitem[Replika()]{replika}
Replika.
\newblock Replika.
\newblock URL \url{https://replika.com/}.

\bibitem[Roller et~al.(2020)Roller, Dinan, Goyal, Ju, Williamson, Liu, Xu, Ott, Shuster, Smith, Boureau, and Weston]{roller2020recipes}
Stephen Roller, Emily Dinan, Naman Goyal, Da~Ju, Mary Williamson, Yinhan Liu, Jing Xu, Myle Ott, Kurt Shuster, Eric~M. Smith, Y-Lan Boureau, and Jason Weston.
\newblock Recipes for building an open-domain chatbot, 2020.

\bibitem[Sanchez et~al.(2023)Sanchez, Fan, Spangher, Levi, Ammanamanchi, and Biderman]{sanchez2023stay}
Guillaume Sanchez, Honglu Fan, Alexander Spangher, Elad Levi, Pawan~Sasanka Ammanamanchi, and Stella Biderman.
\newblock Stay on topic with classifier-free guidance, 2023.

\bibitem[Schulman et~al.(2022)Schulman, Zoph, Kim, Hilton, Menick, Weng, Uribe, Fedus, Metz, Pokorny, Lopes, Zhao, Vijayvergiya, Sigler, Perelman, Voss, Heaton, Parish, Cummings, Nayak, Balcom, Schnurr, Kaftan, Hallacy, Turley, Deutsch, Goel, Ward, Konstantinidis, Zaremba, Ouyang, Bogdonoff, Gross, Medina, Yoo, Lee, Lowe, Mossing, Huizinga, Jiang, Wainwright, Almeida, Lin, Zhang, Xiao, Slama, Bills, Gray, Leike, Pachocki, Tillet, Jain, Brockman, Ryder, Paino, Yuan, Winter, Wang, Bavarian, Babuschkin, Sidor, Kanitscheider, Pavlov, Plappert, Tezak, Jun, Zhuk, Pong, Kaiser, Tworek, Carr, Weng, Agarwal, Cobbe, Kosaraju, Power, Polu, Han, Puri, Jain, Chess, Gibson, Boiko, Parparita, Tootoonchian, Kosic, and Hesse]{chatgpt}
John Schulman, Barret Zoph, Christina Kim, Jacob Hilton, Jacob Menick, Jiayi Weng, Juan Felipe~Ceron Uribe, Liam Fedus, Luke Metz, Michael Pokorny, Rapha~Gontijo Lopes, Shengjia Zhao, Arun Vijayvergiya, Eric Sigler, Adam Perelman, Chelsea Voss, Mike Heaton, Joel Parish, Dave Cummings, Rajeev Nayak, Valerie Balcom, David Schnurr, Tomer Kaftan, Chris Hallacy, Nicholas Turley, Noah Deutsch, Vik Goel, Jonathan Ward, Aris Konstantinidis, Wojciech Zaremba, Long Ouyang, Leonard Bogdonoff, Joshua Gross, David Medina, Sarah Yoo, Teddy Lee, Ryan Lowe, Dan Mossing, Joost Huizinga, Roger Jiang, Carroll Wainwright, Diogo Almeida, Steph Lin, Marvin Zhang, Kai Xiao, Katarina Slama, Steven Bills, Alex Gray, Jan Leike, Jakub Pachocki, Phil Tillet, Shantanu Jain, Greg Brockman, Nick Ryder, Alex Paino, Qiming Yuan, Clemens Winter, Ben Wang, Mo~Bavarian, Igor Babuschkin, Szymon Sidor, Ingmar Kanitscheider, Mikhail Pavlov, Matthias Plappert, Nik Tezak, Heewoo Jun, William Zhuk, Vitchyr Pong, Lukasz Kaiser, Jerry Tworek, Andrew
  Carr, Lilian Weng, Sandhini Agarwal, Karl Cobbe, Vineet Kosaraju, Alethea Power, Stanislas Polu, Jesse Han, Raul Puri, Shawn Jain, Benjamin Chess, Christian Gibson, Oleg Boiko, Emy Parparita, Amin Tootoonchian, Kyle Kosic, and Christopher Hesse.
\newblock Introducing {C}hat{GPT}, 2022.
\newblock URL \url{https://openai.com/blog/chatgpt}.

\bibitem[Shevlane et~al.(2023)Shevlane, Farquhar, Garfinkel, Phuong, Whittlestone, Leung, Kokotajlo, Marchal, Anderljung, Kolt, Ho, Siddarth, Avin, Hawkins, Kim, Gabriel, Bolina, Clark, Bengio, Christiano, and Dafoe]{shevlane2023model}
Toby Shevlane, Sebastian Farquhar, Ben Garfinkel, Mary Phuong, Jess Whittlestone, Jade Leung, Daniel Kokotajlo, Nahema Marchal, Markus Anderljung, Noam Kolt, Lewis Ho, Divya Siddarth, Shahar Avin, Will Hawkins, Been Kim, Iason Gabriel, Vijay Bolina, Jack Clark, Yoshua Bengio, Paul Christiano, and Allan Dafoe.
\newblock Model evaluation for extreme risks, 2023.

\bibitem[Shuster et~al.(2022)Shuster, Xu, Komeili, Ju, Smith, Roller, Ung, Chen, Arora, Lane, Behrooz, Ngan, Poff, Goyal, Szlam, Boureau, Kambadur, and Weston]{shuster2022blenderbot}
Kurt Shuster, Jing Xu, Mojtaba Komeili, Da~Ju, Eric~Michael Smith, Stephen Roller, Megan Ung, Moya Chen, Kushal Arora, Joshua Lane, Morteza Behrooz, William Ngan, Spencer Poff, Naman Goyal, Arthur Szlam, Y-Lan Boureau, Melanie Kambadur, and Jason Weston.
\newblock Blenderbot 3: a deployed conversational agent that continually learns to responsibly engage, 2022.

\bibitem[Skantze(2021)]{turntakingsurvey}
Gabriel Skantze.
\newblock Turn-taking in conversational systems and human-robot interaction: A review.
\newblock \emph{Computer Speech \& Language}, 67:\penalty0 101178, 2021.
\newblock ISSN 0885-2308.
\newblock \doi{https://doi.org/10.1016/j.csl.2020.101178}.
\newblock URL \url{https://www.sciencedirect.com/science/article/pii/S088523082030111X}.

\bibitem[Team et~al.(2024)Team, Mesnard, Hardin, Dadashi, Bhupatiraju, Pathak, Sifre, Rivière, Kale, Love, Tafti, Hussenot, Chowdhery, Roberts, Barua, Botev, Castro-Ros, Slone, Héliou, Tacchetti, Bulanova, Paterson, Tsai, Shahriari, Lan, Choquette-Choo, Crepy, Cer, Ippolito, Reid, Buchatskaya, Ni, Noland, Yan, Tucker, Muraru, Rozhdestvenskiy, Michalewski, Tenney, Grishchenko, Austin, Keeling, Labanowski, Lespiau, Stanway, Brennan, Chen, Ferret, Chiu, Mao-Jones, Lee, Yu, Millican, Sjoesund, Lee, Dixon, Reid, Mikuła, Wirth, Sharman, Chinaev, Thain, Bachem, Chang, Wahltinez, Bailey, Michel, Yotov, Sessa, Chaabouni, Comanescu, Jana, Anil, McIlroy, Liu, Mullins, Smith, Borgeaud, Girgin, Douglas, Pandya, Shakeri, De, Klimenko, Hennigan, Feinberg, Stokowiec, hui Chen, Ahmed, Gong, Warkentin, Peran, Giang, Farabet, Vinyals, Dean, Kavukcuoglu, Hassabis, Ghahramani, Eck, Barral, Pereira, Collins, Joulin, Fiedel, Senter, Andreev, and Kenealy]{gemma}
Gemma Team, Thomas Mesnard, Cassidy Hardin, Robert Dadashi, Surya Bhupatiraju, Shreya Pathak, Laurent Sifre, Morgane Rivière, Mihir~Sanjay Kale, Juliette Love, Pouya Tafti, Léonard Hussenot, Aakanksha Chowdhery, Adam Roberts, Aditya Barua, Alex Botev, Alex Castro-Ros, Ambrose Slone, Amélie Héliou, Andrea Tacchetti, Anna Bulanova, Antonia Paterson, Beth Tsai, Bobak Shahriari, Charline~Le Lan, Christopher~A. Choquette-Choo, Clément Crepy, Daniel Cer, Daphne Ippolito, David Reid, Elena Buchatskaya, Eric Ni, Eric Noland, Geng Yan, George Tucker, George-Christian Muraru, Grigory Rozhdestvenskiy, Henryk Michalewski, Ian Tenney, Ivan Grishchenko, Jacob Austin, James Keeling, Jane Labanowski, Jean-Baptiste Lespiau, Jeff Stanway, Jenny Brennan, Jeremy Chen, Johan Ferret, Justin Chiu, Justin Mao-Jones, Katherine Lee, Kathy Yu, Katie Millican, Lars~Lowe Sjoesund, Lisa Lee, Lucas Dixon, Machel Reid, Maciej Mikuła, Mateo Wirth, Michael Sharman, Nikolai Chinaev, Nithum Thain, Olivier Bachem, Oscar Chang, Oscar
  Wahltinez, Paige Bailey, Paul Michel, Petko Yotov, Pier~Giuseppe Sessa, Rahma Chaabouni, Ramona Comanescu, Reena Jana, Rohan Anil, Ross McIlroy, Ruibo Liu, Ryan Mullins, Samuel~L Smith, Sebastian Borgeaud, Sertan Girgin, Sholto Douglas, Shree Pandya, Siamak Shakeri, Soham De, Ted Klimenko, Tom Hennigan, Vlad Feinberg, Wojciech Stokowiec, Yu~hui Chen, Zafarali Ahmed, Zhitao Gong, Tris Warkentin, Ludovic Peran, Minh Giang, Clément Farabet, Oriol Vinyals, Jeff Dean, Koray Kavukcuoglu, Demis Hassabis, Zoubin Ghahramani, Douglas Eck, Joelle Barral, Fernando Pereira, Eli Collins, Armand Joulin, Noah Fiedel, Evan Senter, Alek Andreev, and Kathleen Kenealy.
\newblock Gemma: Open models based on gemini research and technology, 2024.
\newblock URL \url{https://arxiv.org/abs/2403.08295}.

\bibitem[Team()]{oyez}
Oyez Team.
\newblock About oyez.
\newblock URL \url{https://www.oyez.org/about}.

\bibitem[Thompson et~al.()Thompson, Colebatch, Brown, Rothwell, Day, Obeso, and Marsden]{humanVisualReactTime}
PD~Thompson, JG~Colebatch, P~Brown, JC~Rothwell, BL~Day, JA~Obeso, and CD~Marsden.
\newblock Voluntary stimulus-sensitive jerks and jumps mimicking myoclonus or pathological startle syndromes.
\newblock \doi{10.1002/mds.870070312}.
\newblock URL \url{https://pubmed.ncbi.nlm.nih.gov/1620144/}.

\bibitem[Thoppilan et~al.(2022)Thoppilan, Freitas, Hall, Shazeer, Kulshreshtha, Cheng, Jin, Bos, Baker, Du, Li, Lee, Zheng, Ghafouri, Menegali, Huang, Krikun, Lepikhin, Qin, Chen, Xu, Chen, Roberts, Bosma, Zhao, Zhou, Chang, Krivokon, Rusch, Pickett, Srinivasan, Man, Meier-Hellstern, Morris, Doshi, Santos, Duke, Soraker, Zevenbergen, Prabhakaran, Diaz, Hutchinson, Olson, Molina, Hoffman-John, Lee, Aroyo, Rajakumar, Butryna, Lamm, Kuzmina, Fenton, Cohen, Bernstein, Kurzweil, Aguera-Arcas, Cui, Croak, Chi, and Le]{thoppilan2022lamda}
Romal Thoppilan, Daniel~De Freitas, Jamie Hall, Noam Shazeer, Apoorv Kulshreshtha, Heng-Tze Cheng, Alicia Jin, Taylor Bos, Leslie Baker, Yu~Du, YaGuang Li, Hongrae Lee, Huaixiu~Steven Zheng, Amin Ghafouri, Marcelo Menegali, Yanping Huang, Maxim Krikun, Dmitry Lepikhin, James Qin, Dehao Chen, Yuanzhong Xu, Zhifeng Chen, Adam Roberts, Maarten Bosma, Vincent Zhao, Yanqi Zhou, Chung-Ching Chang, Igor Krivokon, Will Rusch, Marc Pickett, Pranesh Srinivasan, Laichee Man, Kathleen Meier-Hellstern, Meredith~Ringel Morris, Tulsee Doshi, Renelito~Delos Santos, Toju Duke, Johnny Soraker, Ben Zevenbergen, Vinodkumar Prabhakaran, Mark Diaz, Ben Hutchinson, Kristen Olson, Alejandra Molina, Erin Hoffman-John, Josh Lee, Lora Aroyo, Ravi Rajakumar, Alena Butryna, Matthew Lamm, Viktoriya Kuzmina, Joe Fenton, Aaron Cohen, Rachel Bernstein, Ray Kurzweil, Blaise Aguera-Arcas, Claire Cui, Marian Croak, Ed~Chi, and Quoc Le.
\newblock Lamda: Language models for dialog applications, 2022.

\bibitem[Touvron et~al.(2023)Touvron, Martin, Stone, Albert, Almahairi, Babaei, Bashlykov, Batra, Bhargava, Bhosale, Bikel, Blecher, Ferrer, Chen, Cucurull, Esiobu, Fernandes, Fu, Fu, Fuller, Gao, Goswami, Goyal, Hartshorn, Hosseini, Hou, Inan, Kardas, Kerkez, Khabsa, Kloumann, Korenev, Koura, Lachaux, Lavril, Lee, Liskovich, Lu, Mao, Martinet, Mihaylov, Mishra, Molybog, Nie, Poulton, Reizenstein, Rungta, Saladi, Schelten, Silva, Smith, Subramanian, Tan, Tang, Taylor, Williams, Kuan, Xu, Yan, Zarov, Zhang, Fan, Kambadur, Narang, Rodriguez, Stojnic, Edunov, and Scialom]{touvron2023llama}
Hugo Touvron, Louis Martin, Kevin Stone, Peter Albert, Amjad Almahairi, Yasmine Babaei, Nikolay Bashlykov, Soumya Batra, Prajjwal Bhargava, Shruti Bhosale, Dan Bikel, Lukas Blecher, Cristian~Canton Ferrer, Moya Chen, Guillem Cucurull, David Esiobu, Jude Fernandes, Jeremy Fu, Wenyin Fu, Brian Fuller, Cynthia Gao, Vedanuj Goswami, Naman Goyal, Anthony Hartshorn, Saghar Hosseini, Rui Hou, Hakan Inan, Marcin Kardas, Viktor Kerkez, Madian Khabsa, Isabel Kloumann, Artem Korenev, Punit~Singh Koura, Marie-Anne Lachaux, Thibaut Lavril, Jenya Lee, Diana Liskovich, Yinghai Lu, Yuning Mao, Xavier Martinet, Todor Mihaylov, Pushkar Mishra, Igor Molybog, Yixin Nie, Andrew Poulton, Jeremy Reizenstein, Rashi Rungta, Kalyan Saladi, Alan Schelten, Ruan Silva, Eric~Michael Smith, Ranjan Subramanian, Xiaoqing~Ellen Tan, Binh Tang, Ross Taylor, Adina Williams, Jian~Xiang Kuan, Puxin Xu, Zheng Yan, Iliyan Zarov, Yuchen Zhang, Angela Fan, Melanie Kambadur, Sharan Narang, Aurelien Rodriguez, Robert Stojnic, Sergey Edunov, and Thomas
  Scialom.
\newblock Llama 2: Open foundation and fine-tuned chat models, 2023.
\newblock URL \url{https://arxiv.org/abs/2307.09288}.

\bibitem[Turing(1950)]{turing1950computing}
A.~M. Turing.
\newblock Computing machinery and intelligence.
\newblock \emph{Mind}, 59\penalty0 (236):\penalty0 433--460, 1950.
\newblock ISSN 00264423.
\newblock URL \url{http://www.jstor.org/stable/2251299}.

\bibitem[Weizenbaum(1966)]{eliza}
Joseph Weizenbaum.
\newblock Eliza—a computer program for the study of natural language communication between man and machine.
\newblock \emph{Commun. ACM}, 9\penalty0 (1):\penalty0 36–45, jan 1966.
\newblock ISSN 0001-0782.
\newblock \doi{10.1145/365153.365168}.
\newblock URL \url{https://doi.org/10.1145/365153.365168}.

\end{thebibliography}
\bibliographystyle{template}

\appendix

\newpage

\section{Causal Rejection Sampling Algorithm with Speculative Decoding}
\label{app:speculative-algorithm}
{ 
In Algorithm~\ref{alg:speculative-alg}, we present a description of causal rejection sampling including speculative decoding. For simplicity, we describe the speculative rejection sampling as if it rejects or accepts an entire event, but in implementations where events are composed of multiple tokens, the acceptance/rejection acts in finer granularity on tokens (so a prefix in a speculated event can be accepted, and only the rest has to be resampled).

Note that in order to maintain the validity of the rejection sampling, we must condition the draft distribution on $t_i \geq T$, because if $\hat{t}_i$ had been $< T$ it would have already been finalized and we would not be considering it for rejection sampling. Renormalizing this correctly in timestamps consisting of multiple tokens requires some finesse. For our instant messenger case, this is further complicated by the fact that as interruptions come in, the same timestamp in a draft message may change format. For example, after a message planned for \texttt{:02;17.8}, a generation may plan for \texttt{:03;52.0}, but after an interruption at \texttt{:03;24.7}, the draft must be reinterpreted to have a timestamp of \texttt{;52.0}). We expect that speculation is not worth the implementation burden unless you are using a simple custom vocabulary for timestamps or pressing up against performance limits. In our experiments performance was adequate without it.
}

\begin{figure}[h]
\centering
\vspace{-4mm}
\begin{minipage}{0.85\linewidth}
\begin{algorithm}[H]
\caption{Causal rejection sampling with speculation}\label{alg:speculative-alg}
\begin{algorithmic}
\fontsize{8}{9}\selectfont
\State $e \gets []$ \Comment{event history}
\State $i \gets 0$ \Comment{current event index}
\State $c \gets (\varnothing, \varnothing, \varnothing)$ \Comment{candidate for the next message}
\State $r \gets$ false \Comment{whether a candidate was just rejected}
\While {true}
\State $i \gets i + 1$
\Try
\State $(\hat{t}, \hat{s}, \hat{m}) \gets c$
\If{$\hat{t}$ is $\varnothing$}
\State $P \overset{\Delta}{=} p(e_i | e_1, ..., e_{i-1})$
\If{$r$}
\State $Q \overset{\Delta}{=} p(e_{i-1} | e_1, ..., e_{i-2}, t_{i-1} \geq T)$
\State $c \gets (\hat{t}, \hat{s}, \hat{m}) \sim norm(max(0,P-Q))$
\State $r \gets$ false
\Else
\State $c \gets (\hat{t}, \hat{s}, \hat{m}) \sim P$
\EndIf
\EndIf
\State \textbf{wait} until $\hat{t}$
\State $t_{cur} \gets \hat{t}$
\If{$\hat{s}$ is $S$}
\State $c \gets (\varnothing, \varnothing, \varnothing)$
\State $i \gets i-1$
\State \textbf{continue}
\EndIf
\EndTry
\Catch{user input $(T, S, M)$}
\State $e_i \gets (T, S, M)$
\State $t_{cur} \gets T$
\State $(\hat{t}, \hat{s}, \hat{m}) \gets c$
\If {$\hat{t} + t_{react} < T$}
\If {$\hat{s} \neq S$}
\State $P \overset{\Delta}{=} p(e_{i+1}=c | e_1, ..., e_i=(T, S, M))$
\State $Q \overset{\Delta}{=} p(e_i=c | e_1, ..., e_{i-1}, t_i \geq T)$
\If{$Q \leq P$}
\State \textbf{continue}
\Else
\If{$u \sim U[0, 1] > 1 - \frac{P}{Q}$}
\State \textbf{continue}
\Else
\State $r \gets$ true
\EndIf
\EndIf
\EndIf
\State $c \gets (\varnothing, \varnothing, \varnothing)$
\EndIf
\State \textbf{continue}
\EndCatch
\State $e_i \gets c$
\State \textbf{emit} $c$
\State $c \gets (\varnothing, \varnothing, \varnothing)$
\EndWhile
\end{algorithmic}
\end{algorithm}
\end{minipage}
\end{figure}

\newpage

\section{Causal Rejection Sampling Algorithm with Retconning (without Speculative Decoding)}
\label{app:spoken-algorithm}

In Algorithm~\ref{alg:retcon-alg}, we present a description of causal rejection sampling with retconning, which supports streaming word-level ASR input where previous inputs may change retroactively given new context. For simplicity, we do not include speculative decoding, though it could also be applied upon retconning.

\begin{figure}[h]
\centering
\begin{minipage}{\linewidth}
\begin{algorithm}[H]
\caption{Causal rejection sampling with retconning (without speculative decoding)}\label{alg:retcon-alg}
\begin{algorithmic}
\State $i \gets 0$ \Comment{current event index}
\State $e \gets []$ \Comment{event history}
\State $c \gets (\varnothing, \varnothing, \varnothing)$ \Comment{candidate for the next message}
\While {true}
\State $i \gets i + 1$
\Try
\State $(\hat{t}, \hat{s}, \hat{m}) \gets c$
\If{$\hat{t}$ is $\varnothing$}
\State $c \gets (\hat{t}, \hat{s}, \hat{m}) \sim p(e_i | e_1, ..., e_{i-1})$
\EndIf
\State \textbf{wait} until $\hat{t}$
\State $t_{cur} \gets \hat{t}$
\If{$\hat{s}$ is $S$}
\State $c \gets (\varnothing, \varnothing, \varnothing)$
\State $i \gets i-1$
\State \textbf{continue}
\EndIf
\EndTry
\Catch{user input $(T, S, M)$}
\State $e_i \gets (T, S, M)$
\State $t_{cur} \gets T$
\State $(\hat{t}, \hat{s}, \hat{m}) \gets c$
\If {$\hat{s} = S$ or $\hat{t} + t_{react} < T$}
\State $c \gets (\varnothing, \varnothing, \varnothing)$
\EndIf
\State \textbf{continue}
\EndCatch
\Catch{user retcon $j, (T, S, M), t_{cur}$}
\State $e_j \gets (T, S, M)$
\State $(\hat{t}, \hat{s}, \hat{m}) \gets c$
\If {$\hat{s} = S$ or $\hat{t} + t_{react} < t_{current}$}
\State $c \gets (\varnothing, \varnothing, \varnothing)$
\EndIf
\State $i \gets i-1$
\State \textbf{continue}
\EndCatch
\State $e_i \gets c$
\State \textbf{emit} $c$
\State $c \gets (\varnothing, \varnothing, \varnothing)$
\EndWhile
\end{algorithmic}
\end{algorithm}
\end{minipage}
\end{figure}

\newpage
\section{Finetuning Details}
\label{app:finetuning}

We finetune Pythia 160M, Pythia 1.2B, Pythia 12B, Gemma 2B, and Llama 2 7B. We finetune for several epochs with learning rate $10^{-5}$ and batch size 512. We use early stopping on validation loss (computed once per epoch) with a minimum delta of 0.01 and a patience of 3.

\section{In-Context Learning Details}
\label{app:in-context-learning}

For our ICL experiments using GPT and Claude, we use the following system prompts and decode with default sampling parameters.

\textbf{Instant messenger dialogues:}
\begin{verbatim}
Your job is to continue instant messenger conversations between two individuals
inspired by a partial transcript of their chat history. Your generated 
conversations must be new (i.e., they should not appear in whole or in part in 
the transcript), but they should be stylistically and factually consistent with 
the transcript. You must preserve the characterization of both individuals as much 
as possible. DO NOT include anything in your response except a continuation of the 
provided conversation transcript in the same format as the chat transcript. Output 
nothing else, either before or after the continuation.

Each message in the chat transcript is formatted as follows:

<timestamp><user><message><delimiter>

A full transcript consists of many messages in this format concatenated 
together without any whitespace. A sample message is given below:

2023June04Su+01:53;42.7Anow that you mention it-------+

The <timestamp> field includes the year (e.g. "2023"), the month (e.g. "June") 
the date (e.g. "04"), one or two letters denoting the weekday (e.g. "Su") , and 
then the time in UTC (e.g. "01:53;42.7"), all concatenated without spaces in 
that order. If any part of the timestamp is the same as in the previous message, 
it is omitted to save space.

The <user> field is either "A" or "B".

The <message> field is an arbitrary string (in this case "now that you mention 
it").

<delimiter> is always "-------+".
\end{verbatim}

\textbf{Spoken conversation:}

\begin{verbatim}
Your job is to continue timed speech transcripts of Supreme Court arguments. 
Your generated transcript completions must be new (i.e., they should not appear 
in whole or in part in the transcript), but they should be stylistically and 
factually consistent with the transcript. You must preserve the characterization 
of the speakers as much as possible. DO NOT include anything in your response 
except a continuation of the provided transcript in the same format as the 
transcript. Output nothing else, either before or after the continuation.

Each word of the transcript is formatted as follows:

<timestamp><speaker><word>

A full transcript consists of many words in this format concatenated together 
with one space in between. A sample message is given below:

131Gwe'll

The <timestamp> field is three digits denoting the seconds place of the time, 
the decisecond, and the centisecond.

The <speaker> field is a single capital letter.

The <word> field is a single word.

<transcript>
\end{verbatim}

\section{More Quantitative Results}
\label{app:more-quantitative}

For the instant messenger human ratings: offline ratings are averaged over 20 examples per model and online ratings are averaged over 5. For the spoken conversations human ratings: offline ratings are averaged over 10 examples per model and online ratings are averaged over 5.

\begin{figure}[t]
\includegraphics[width=\textwidth]{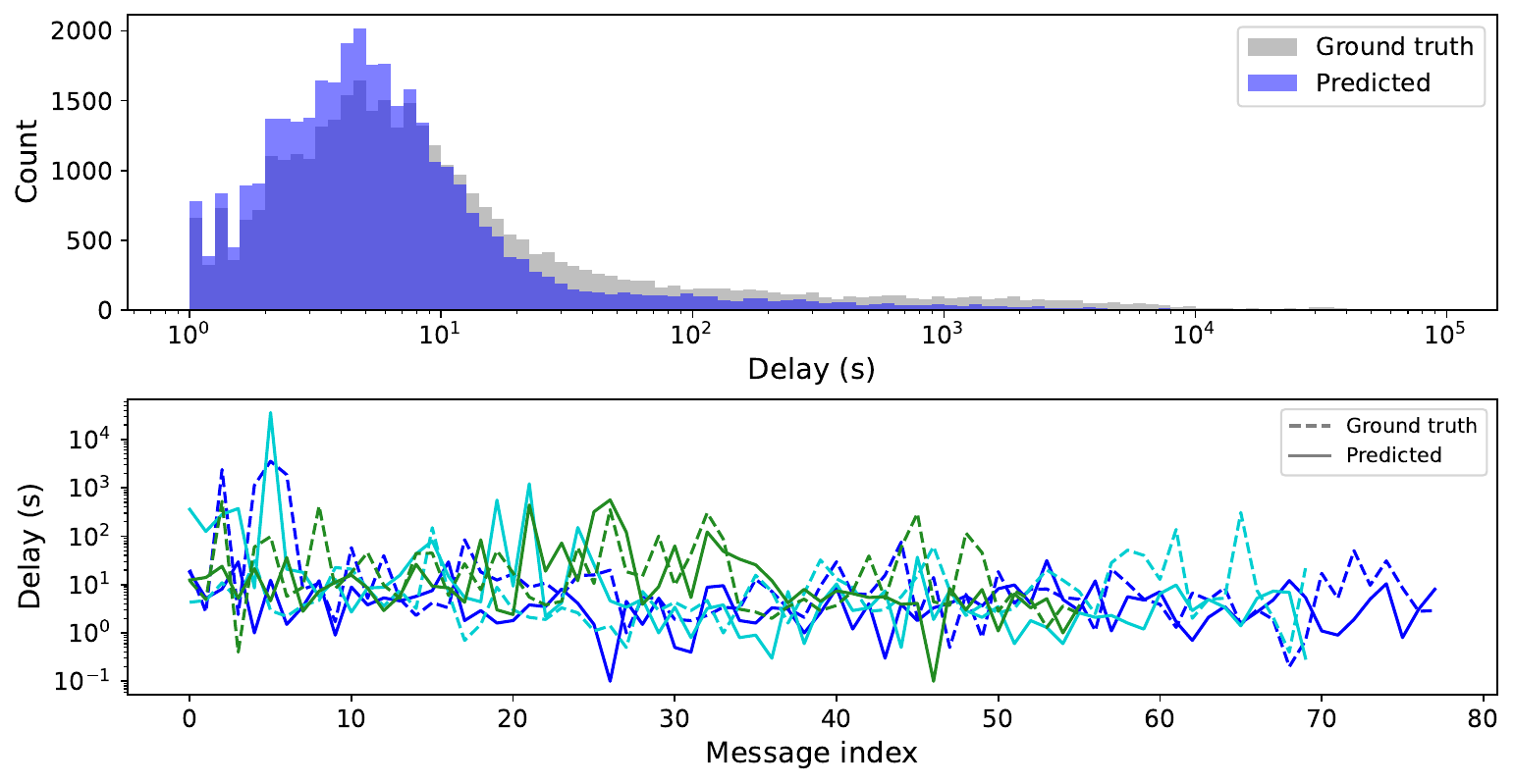}
\caption{\textbf{Conversations generated by fine-tuned language models exhibit realistic message timings.} \textit{Top:} Log-binned histogram of the delays (in seconds) between successive messages in 512 independent 1000-token conversations generated unconditionally by fine-tuned Llama 2 7B (temperature 1, top-p=0.95~\citep{nucleus_sampling}), compared to delays in a corresponding chunk of consecutive ground-truth messages of the same size sampled at random from the same month and year as the simulated ones. Mean conversation length is 73 messages. The empirical distributions are very similar (25-bin Kullback–Leibler divergence = 0.005), attributable to nucleus sampling. \textit{Bottom}: Consecutive message delays for continuations of three randomly selected message history prefixes, ground truth (dotted) vs. predicted (solid). We do not expect these to perfectly match due to irreducible entropy, but the resemblance in trajectory shows that the model is not just learning first-order statistics.}
\label{fig:timestamps}
\end{figure}

See Figure~\ref{fig:timestamps} for experiments comparing the distribution of predicted timestamps to the ground truth distribution.

\section{Qualitative Examples}
\label{app:qualitative-examples}
See below for examples of ground truth and generated conversations, for both the instant messenger dialogue and spoken conversation case studies.

See Figure~\ref{fig:gt-messenger-example} for a ground truth example, Figure~\ref{fig:pythia-160m-messenger-example} for a Pythia 160M example, Figure~\ref{fig:pythia-1p4b-messenger-example} for a Pythia 1.4B example, Figure~\ref{fig:pythia-12b-messenger-example} for a Pythia 12B example, Figure~\ref{fig:gemma-messenger-example} for a Gemma 2B example, 
Figure~\ref{fig:llama-messenger-example} for a Llama 2 7B example, 
Figure~\ref{fig:claude3-messenger-example} for a Claude 3 Sonnet example, and Figure~\ref{fig:gpt4-messenger-example} for a GPT-4 Turbo example.

See Figure~\ref{fig:gt-spoken-example} for a ground truth example, Figure~\ref{fig:pythia-160m-spoken-example} for a Pythia 160M example, Figure~\ref{fig:pythia-1p4b-spoken-example} for a Pythia 1.4B example, Figure~\ref{fig:pythia-12b-spoken-example} for a Pythia 12B example, Figure~\ref{fig:gemma-spoken-example} for a Gemma 2B example, 
Figure~\ref{fig:llama-spoken-example} for a Llama 2 7B example, 
Figure~\ref{fig:claude3-spoken-example} for a Claude 3 Sonnet example, and Figure~\ref{fig:gpt4-spoken-example} for a GPT-4 Turbo example. \\

\newpage

\begin{figure}[h]
\centering
\begin{minipage}{0.875\textwidth}
\begin{verbatim}
:55;50.0AI think chatbot + speechbot could be a nice paper together
:56;16.6Athe contribution is real-time language modeling with
         timestamp decoding

                          ;19.1Bthe worst thing about this paper
               ;25.0Bis that some combination of the paper itself
                               ;28.5Band the cluster difficulties
   ;38.4Bmakes it feel like I'm just endlessly repeating the same
                                                             task

;43.8Ayes

\end{verbatim}
\end{minipage}
\caption{\textbf{Ground truth instant messenger example.}}
\label{fig:gt-messenger-example}
\end{figure}

\newpage
\begin{figure}[h]
\centering
\begin{minipage}{0.875\textwidth}
\begin{verbatim}

:11;35.5Ahttps://www.[redacted]/watch?v=jG-IHVtU-Y

                          :48;02.8Bhttps://[redacted]/C4vNE7CdK0

;05.5Ahttps://[redacted]/2/?l=ms/4401127580322316

                                                    :55;18.7Bsad

;29.6Ahttps://www.[redacted]/videos/202816392889398/
;34.0Amy comment is so rikt'd
:56;01.5Ahttps://[redacted]/status/1716488523789759696
;05.1Aalso those responses
;08.8Alol
;15.8Athe adults are not actually that bad
;32.3AI mean
;32.8Athey sent humans attached
;35.2Ain this case
;37.8Athe two houses are just some random images of babies tomoe
:58;06.2Awould look like a tree in hans graham's apple courtyard
:59;23.4Athey're just natural trees
;38.3Aare they scientific creation?
+22:10;08.4Athe place where black hole black holes go from is not
            really an underground world
;37.9Aand they're everywhere
:12;35.4Ayes

                                      ;40.8Bthey're moving fingers  
                                      
\end{verbatim}
\end{minipage}
\caption{\textbf{Pythia 160M instant messenger example.}}
\label{fig:pythia-160m-messenger-example}
\end{figure}

\newpage
\begin{figure}[h]
\centering
\begin{minipage}{0.875\textwidth}
\begin{verbatim}
                                                       ;07.5Blol
                                                   ;17.6Bso good
                       ;22.3B"if all this other stuff were true"
          ;27.0B"you probably should not have lied to the world"
             :09;24.7BI need to stop being so cynical about this
                                              ;26.5Bit's so good
      ;40.0BI'll only be cynical if I have to justify it somehow

:14;33.8Ayes
;53.0AI can't imagine why you would have that
      in your head though

                                      :15;07.9Bit's pretty funny
                      ;11.4BI told you how my friend who is like
                                             ;13.81.4BM students
                                                   ;16.5Bgets in
                                ;19.0Band asks me to come to his
                                          ;22.1BI am a sophomore
                                                 ;24.2Bit's like
                                         ;26.9Bit's the same age
                       ;29.9Bbut he's in a more selective school

;47.5Ayes I don't understand how people are this cynical

                ;53.8Bhow did you become a social critic so late
                       :16;21.4Band have this incredible network
                                 ;23.9BI am in that category too
   ;31.7Bthere's probably a social media layer under the network
       ;36.8Bwhere people who are in the school in a certain way
                              ;39.3Bare probably very good at it
    ;44.7Blike maybe a few years ago when I first found my niche
                              I probably had to work really hard
                                          ;46.6Bto get attention
                  ;56.0Bbut you kind of need to do it constantly

.5Ayes that's what I was thinking of when I said the other day

                                                 ;57.4Bit's like  

\end{verbatim}
\end{minipage}
\caption{\textbf{Pythia 1.4B instant messenger example.}}
\label{fig:pythia-1p4b-messenger-example}
\end{figure}

\newpage
\begin{figure}[h]
\centering
\begin{minipage}{0.875\textwidth}
\begin{verbatim}
;44.0Ait's going to talk about earth
:48;17.4Alol
;18.6Ait did
:52;40.7AI can't believe how good it is
:53;05.8Ait can apparently hear a person's breathing and then say
         what they think is the most likely reason for that
;10.1Alike
;18.7A"I'm guessing it's the result of you exhaling"
;19.4A"lol"
;22.4AI didn't even understand it

     :57;00.1Bis it so good at understanding what is being asked
  :58;05.8Bit can tell what people are talking about pretty well

;26.1Athat is impressive
;33.3Abut what do they mean by that

           ;34.8Bit can anticipate what people will ask it about
                                       ;39.6Band guess correctly

;44.7Athat's kind of an illusion I think
;51.0Aespecially at test time

                                   :59;02.8Bit's not an illusion
                                         ;04.2Bit's a prediction
                                 ;17.2Bthis one is from the book
                            ;26.0Bit's just really fast and deep
             ;36.8Bmaybe it's really good at interpreting images
      ;45.3BI think it has some kind of image memory of whatever
                              it saw the person doing previously

:59;24.9Ayes but it was probably trained with that data at least
;29.4Abut maybe they are good at understanding like
;32.4A"which object in the picture"
;33.9Arather than

                            .9Bit probably doesn't remember like
                                ;38.5B"I've seen the sun before"
   ;44.3B"let me predict that you'll say 'it looks bright on the
                                                    left side"'"
                                                 ;45.3Byes it is
                                                  
\end{verbatim}
\end{minipage}
\caption{\textbf{Pythia 12B instant messenger example.}}
\label{fig:pythia-12b-messenger-example}
\end{figure}

\newpage
\begin{figure}[h]
\centering
\begin{minipage}{0.875\textwidth}
\begin{verbatim}
:19;01.7Ait's hard to say

                               ;08.8Bnot as long as you do stuff
                                     ;11.9Byou can still do lots

;16.4Ayes

                                    ;21.0BI get exhausted though
                        ;35.3BI'm so tired after a day like that

;44.1Alol

                      ;58.7Bso you could definitely do more work

:20;03.0AI get more motivated to start
;09.1Abut then I'm constantly worried

                               ;14.0Bthat's probably a positive!

;17.9AI'll get bored

   ;25.4Bmaybe less productive than someone without that feeling

;28.0Aso I have to just do something

                           ;30.5Bif you need to have things done

;34.3Ayes

                               ;35.2Bthen have more things to do

;37.5Athen why not work

                        ;42.4BI literally can't work on anything
                                                  ;43.6Bfor days

.9Aif you're not already doing work
;45.8Ayou need to stop thinking
;51.5Aand just do it

                   :21;24.1Blike this week has been unproductive 

\end{verbatim}
\end{minipage}
\caption{\textbf{Gemma 2B instant messenger example.}}
\label{fig:gemma-messenger-example}
\end{figure}

\newpage
\begin{figure}[h]
\centering
\begin{minipage}{0.875\textwidth}
\begin{verbatim}
                                      ;41.8Bit's obviously false
                      ;47.3Bthe fact that it's true is the proof
                                              ;56.0Bit's so good
                                                   ;58.1Bso deep
                                              :50;01.9Bthe irony
                                        ;25.2Bbut then it's like
            ;32.3Bwhat are the implications of the obvious truth
   ;58.7Bwho is going to write a textbook called "complexity and
                                                   universality"
    :51;17.6Bthere's no way any human in this class would get it
                ;46.6BI guess it's just something professors say
               :52;07.1Bjust one page in every 300 page textbook
                          ;15.4Band like three times in the rest
                           ;18.9Bwhen they talk about complexity

:53;00.5Amy friend sent me a list of all her classes so far
;04.4Aand they're all so interesting
;09.7Alike one was a class on china
;13.4Aand another was the music of west africa
;22.2Athat reminds me
;25.8Awhat happened to your internships

                             ;33.0BI've just been sitting around
                          ;40.2Bwaiting for interviews to happen
               :54;02.2BI got a call about a different one today

;03.4Af

                         ;09.6Bbut I don't want to be doing that

;13.0Ayes

               ;17.4BI'm waiting for the ML interviews to happen
            ;25.2Bhopefully the university one will come through
                       ;29.1Bsince that's the most promising one

;56.9Awhat else were the other two 
\end{verbatim}
\end{minipage}
\caption{\textbf{Llama 2 7B instant messenger example.}}
\label{fig:llama-messenger-example}
\end{figure}

\newpage
\begin{figure}[h]
\centering
\begin{minipage}{0.875\textwidth}
\begin{verbatim}
  :41;02.7Bj'm surprised my senses aren't triggering too quickly
 ;24.4Bit feels like you'll have to spend half an hour breathing

:43;00.5Alol I see the pop quiz early tomorrow

               :44;34.2Bso I can only focus on writing the essay

;37.7Athen I'll meet you at 12:45
;41.2Aprobably I'll meet you anyway
;43.4Aand I have an appetite for reading your short stories
     one day

                                                     :46;05.3Bhm

:47;58.4AI just finished this letter

 :49;02.4Bthat's just the end of today. I just want to sit still

;46.0Aand read it

                                ;47.2BI have an expectation that
                          ;49.3Bit's going to be fun to write it
                              ;52.9Bbut I don't want to write it
                                                    :50;11.0Bsmh
                                :51;49.1Bit's supposed to be rpg

.8Abut whoops
;52.0AI'm so anxious to finish it

;53.6Bbut then I'll just need to study about word choice at least 

\end{verbatim}
\end{minipage}
\caption{\textbf{Claude 3 Sonnet instant messenger example.}}
\label{fig:claude3-messenger-example}
\end{figure}

\newpage
\begin{figure}[h]
\centering
\begin{minipage}{0.875\textwidth}
\begin{verbatim}
;22;33.5APractically inviting the memes at this point

           ;23;45.9Bcan't believe the world we live in sometimes

;27;14.2Aright?
;29;00.7Aevery day feels like a satirical novel

                             ;31;52.3Bexcept it's painfully real

;35;06.8Aexactly
;38;29.5Aanyway did you finish that analysis?

                                     ;40;03.1Byeah, nearly there
                        ;42;15.6Bgot sidetracked by the news tbh

;45;04.9Aunderstandable
;47;11.3Asend it over when you're done?

                                                ;49;22.8Bwill do
                   ;51;30.9Bneed a distraction from this madness

;53;45.2Asame here
;55;57.9Alet's grab coffee later, need to vent

\end{verbatim}
\end{minipage}
\caption{\textbf{GPT-4 Turbo instant messenger example.}}
\label{fig:gpt4-messenger-example}
\end{figure}

\newpage 
\begin{figure}[h]
\centering
\begin{minipage}{0.875\textwidth}
\begin{verbatim}
229Aany
259Aaction
293Aunder
395Acould
411Abe
425Abrought
457Aincluding
497Aa
505Aquitam
561Aaction
653Aor
683Aincluding
723Aa
729Aretaliation
815Aaction
907Abut
935Athat
961Areading
001Ais
017Anot
043Aessential
105Ato
214Athe
226Aposition
284Athat
298Awe
316Aadvocate
368Ahere
                                                          422Cso
                                                        442Cthat
                                                       532Ccould
                                                     580Cexclude
                                                        746Ceven
                                                           768Ca
                                                 776Cretaliation
                                                      882Caction
                                                          996Cby
                                                          020Can
                                                  032Cindividual
                                                         122Cwho
                                                     199Cdoesn't
                                                        241Cknow
                                                        291Cwhen
                                                         323Cthe
                                                        427Cwhen
                                                         449Cthe
                                                  469Cindividual
                                                         631Cthe
                                                    643Cofficial
                                                          685Cof
                                                         693Cthe
                                                      707Cunited
                                                      747Cstates
                                                       837Cfound
                                                         867Cout
                                                       881Cabout
                                                       907Cthose
                                                       935Cfacts

\end{verbatim}
\end{minipage}
\caption{\textbf{Ground truth spoken conversation example.}}
\label{fig:gt-spoken-example}
\end{figure}

\newpage
\begin{figure}[h]
\centering
\begin{minipage}{0.875\textwidth}
\begin{verbatim}
751Awhite
799Athat
821Ato
837Ayour
853Aoffice
903Ais
999Avery
949Alikely
057Aunder
079Athe
095Aoriginal
155Achapter
209Aso
211Aif
251Ayour
263Amajority
321Acompels
443Athe
456Aattaches
495Athe
511Aperson
553Ais
587Ato
627Amurder
999Aas
039Aby
049Athe
063Aattaches
127Ais
141Athe
171Awhite
222Aman
247Ato
271Ayou
285Aand
299Amays
327Ato
361Ame
375Athe
407Aperson
435Aoffering
475Ainformation
559Afor
565Awork
625Awas
665Athe
681Atype
697Aof
707Aan
719Aprogressive
853Aexpression
963Aof
981Aspeed
989Ashifts
061Ado
081Athat
107Aoccur

\end{verbatim}
\end{minipage}
\caption{\textbf{Pythia 160M spoken conversation example.}}
\label{fig:pythia-160m-spoken-example}
\end{figure}

\newpage
\begin{figure}[h]
\centering
\begin{minipage}{0.875\textwidth}
\begin{verbatim}
719Ayork
745Alaw
773Asay
805Ayou
827Acan't
851Ado
875Aanything
975Ai
983Asuppose
031Ayou
059Aknow
085Ahow
099Ato
109Ado
121Asomething
165Aover
180Athere
                        223Ji
                        229Jdon't
                                                        239Gknow
                                                      255Gthat's
                                                       279Gwhere
                                                         299Gi'm
                                                     319Ggetting
                                                        347Gfrom
                                                        365Gyour
                                                      391Ganswer
                                                         435Gthe
                                                    449Gquestion
                                                          487Gis
                                                         501Gwhy
                                                       535Gisn't
                                                        567Gthat
                                                       585Ggoing
                                                          611Gto
                                                          623Gbe
                                                        637Gdone
                        829Jwell
                        851Jthat's
                        873Jtrue
                        887Jbut
                        933Jone
                        945Jway
                        963Jto
                        977Jdo
                        001Jit
                        019Jis
                        033Jto
                        049Jget
                        073Jthe
                        089Jnotice
                        127Jto
                        139Jthe
                        155Jbank

\end{verbatim}
\end{minipage}
\caption{\textbf{Pythia 1.4B spoken conversation example.}}
\label{fig:pythia-1p4b-spoken-example}
\end{figure}

\newpage
\begin{figure}[h]
\centering
\begin{minipage}{0.875\textwidth}
\begin{verbatim}
312Aflames
407Agoing
469Athere
488Awasn't
515Aa
517Afire
550Ajust
566Agoing
595Athat
613Away
650Athen
670Ait's
684Aan
694Ainvalid
740Asearch
770Aright
                                                        788Fthat
                                                       800Fwould
                                                          816Fbe
                                                          832Fan
                                                     842Finvalid
                                                      870Fsearch
                                                          918Fif
                                                       932Fthere
                                                         948Fwas
                                                       964Fnever
                                                           988Fa
                                                        996Ffire
                                                          036Fat
                                                         042Fthe
                                                       050Fhouse
                                                          076Fif
                                                     084Fthere's
                                                       108Fnever
                                                           140Fa
                                                        148Ffire
                                                          170Fat
                                                         180Fthe
                                                       190Fhouse
                                                         224Fnow
                                                     240Fjustice
                                                     270Fstevens
                                                       300Fasked
                                                         316Fthe
                                                    326Fquestion
                                                         360Fthe
                                                       374Ffacts
                                                         406Fare
                                                        458Fwell
                                                         478Fmay
                                                         498Fnot
                                                        522Fhave
                                                        540Fbeen
                                                           560Fa
                                                        566Ffire
                                                          584Fin
                                                         592Fthe
                                                       604Fhouse

\end{verbatim}
\end{minipage}
\caption{\textbf{Pythia 12B spoken conversation example.}}
\label{fig:pythia-12b-spoken-example}
\end{figure}

\newpage
\begin{figure}[h]
\centering
\begin{minipage}{0.875\textwidth}
\begin{verbatim}
372Abut
400Ahow
430Ado
444Awe
460Aknow
482Awhen
498Athere's
524Aa
532Asecond
568Astep
606Arequired
654Ai
668Athink
692Athat
708Athe
720Areason
758Athat
774Awisconsin
834Afailed
864Ato
882Ado
902Ait
920Ais
932Abecause
962Aof
976Ajustice
016Ascalia's
072Aconcern
134Aabout
156Athe
168Aburdens
218Aon
230Adue
250Aprocess
                                                        352Bwell
                                                      366Bthat's
                                                          390Bin
                                                        402Bpart
                                                           426Ba
                                                     434Bproblem
                                                          460Bwe
                                                        472Bhave
                                                        504Bwith
                                                         516Bthe
                                                  530Brationally
                                                     576Brelated
                                                        614Btest
                                                       644Bwhich
                                                          666Bis
                                                        684Bthat
                                                          762Bif
                                                           782Bi
                                                         798Bput
                                                           820Ba
                                                830Bhypothetical
                                                          900Bon
                                                         908Bthe
                                                       918Btable 
                                                        
\end{verbatim}
\end{minipage}
\caption{\textbf{Gemma 2B spoken conversation example.}}
\label{fig:gemma-spoken-example}
\end{figure}

\newpage
\begin{figure}[h]
\centering
\begin{minipage}{0.875\textwidth}
\begin{verbatim}
096Afor
110Aexample
170Athat
188Acame
212Ainto
232Aevidence
274Athat
304Ahe
318Ahad
338Aa
346Alot
362Aof
370Adifferent
398Adrivers
442Alicenses
520Athere
538Awas
558Aevidence
590Afrom
608Athe
618Amaricopa
674Acounty
714Aattorney's
764Aoffice
790Athat
804Ahe
814Ahad
844Anine
876Aprevious
934Afelony
976Aconvictions
050Ai'm
062Asorry
096Aeight
112Aprevious
160Afelony
198Aconvictions
297Athere
312Awas
339Aevidence
373Athat
397Ahe
415Ahad
481Ahad
497Aan
507Aextensive
561Afraud
595Abankruptcy
655Acase
709Abefore

\end{verbatim}
\end{minipage}
\caption{\textbf{Llama 2 7B spoken conversation example.}}
\label{fig:llama-spoken-example}
\end{figure}

\newpage
\begin{figure}[h]
\centering
\begin{minipage}{0.875\textwidth}
\begin{verbatim}
131Awe'll
563Acounty
676Asoil
                718Gand
                732Gwater
                                766Hconservation
                                846Hdistrict
                                972Hversus
                                013Hthe
                                027Hunited
                                071Hstates
                                007Hmr
                                021Hbrowning
                                                          217Bmr
                                                       245Bchief
                                                     267Bjustice
                                                         331Band
                                                         343Bmay
                                                          359Bit
                                                      369Bplease
                                                         405Bthe
                                                       417Bcourt
                                                         503Bthe
                                                       527Bissue
                                                          565Bin
                                                        575Bthis
                                                        603Bcase
                                                          697Bis
                                                     713Bwhether
                                                    747Bcongress
                                                   831Bexpressly
                                                    901Bprovided
                                                         959Bfor
                                                           977Ba
                                                 985Blimitations
                                                      061Bperiod
                                                         103Bfor
                                                 125Bretaliatory
                                                   199Bdischarge
                                                      269Baction
                                                       325Bunder
                                                         343Bthe
                                                     353Bfederal
                                                       385Bfalse
                                                      421Bclaims
                                                         469Bact
                                                         551Bthe
                                                    567Bsix-year
                                                  613Blimitation
                                                      679Bperiod
                                                         764Bset
                                                         792Bout
                                                          873Bin
                                                     897Bsection
                                                       002B3731b

\end{verbatim}
\end{minipage}
\caption{\textbf{Claude 3 Sonnet spoken conversation example.}}
\label{fig:claude3-spoken-example}
\end{figure}

\newpage
\begin{figure}[h]
\centering
\begin{minipage}{0.875\textwidth}
\begin{verbatim}
464Agive
480Aa
492Awritten
508Astatement
552Awithout
570Athe
582Apresence
609Aof
617Aan
625Aattorney.
                                                      682CThat's
                                                      700Cclear,
                                                         716Cbut
                                                         728Cthe
                                                        736Cfact
                                                        744Cthat
                                                          752Che
                                                         756Cwas
                                                     760Cwilling
                                                          776Cto
                                                       780Cspeak
                                                      794Corally
                                                     823Cwithout
                                                         841Cone
                                                     857Cdoesn't
                                                 865Cnecessarily
                                                        925Cmean
                                                          945Che
                                                  957Cunderstood
                                                         979Cthe
                                                003Cimplications
                                                      055Cfully.
                        102GThat
                        118Gis,
                        132Gdid
                        146Ghe
                        160Gunderstand
                        188Gthat
                        200Gan
                        212Goral
                        226Gstatement
                        262Gcould
                        274Gstill
                        286Gbe
                        298Gused
                        314Gagainst
                        330Ghim
                        344Gin
                        352Ga
                        364Gcourt
                        376Gof
                        384Glaw
                        408Gjust
                        420Gas
                        432Geffectively
                        472Gas
                        486Ga
                        494Gwritten
                        516Gone?
\end{verbatim}
\end{minipage}
\caption{\textbf{GPT-4 Turbo spoken conversation example.}}
\label{fig:gpt4-spoken-example}
\end{figure}

\end{document}